\documentclass{article}
\usepackage{ijcai26}

\usepackage{times}
\usepackage{soul}
\usepackage{url}
\usepackage[hidelinks]{hyperref}
\usepackage[utf8]{inputenc}
\usepackage[small]{caption}
\usepackage{graphicx}
\usepackage{amsmath}
\usepackage{amssymb}
\usepackage{amsthm}
\usepackage{booktabs}
\usepackage{algorithm}
\usepackage{algorithmic}
\usepackage[switch]{lineno}

\usepackage{multirow}
\usepackage{makecell}
\usepackage{float}
\usepackage{colortbl}
\usepackage{xcolor}
\usepackage[table]{xcolor}
\definecolor{lightblue}{RGB}{235, 245, 255}  
\usepackage[most]{tcolorbox}
\usepackage{enumitem}
\usepackage{placeins}
\usepackage{url}

\title{Pedagogical Alignment for Vision-Language-Action Models:\\ A Comprehensive Framework for Data, Architecture, and Evaluation in Education}

\author{
Unggi Lee$^{1\dagger}$ \and
Jahyun Jeong$^{2\dagger}$\and
Sunyoung Shin$^{2\dagger}$\and
Haeun Park$^{3\dagger}$ \and
Jeongsu Moon$^1$ \\
Youngchang Song$^1$ \and
Jaechang Shim$^1$ \and
JaeHwan Lee$^1$ \and
Yunju Noh$^1$ \and
Seungwon Choi$^1$ \\
Ahhyun Kim$^1$ \and
TaeHyeon Kim$^1$ \and
Kyungtae Joo$^1$ \and
Taeyeong Kim$^{1\dagger}$  \And
Gyeonggeon Lee$^{4\dagger}$ 
\\
\affiliations
$^1$Chosun University $^2$Seoul National University\\ $^3$Korea Institute for Curriculum and Evaluation $^4$Nanyang Technological University \\
\textbf{Corresponding Authors (†, contact emails):} codingchild@korea.ac.kr, gyeonggeon.lee@nie.edu.sg
}

\begin{document}

\maketitle

\begin{abstract}

Science demonstrations are important for effective STEM education, yet teachers face challenges in conducting them safely and consistently across multiple occasions, where robotics can be helpful. However, current Vision-Language-Action (VLA) models require substantial computational resources and sacrifice language generation capabilities to maximize efficiency, making them unsuitable for resource-constrained educational settings that require interpretable, explanation-generating systems. We present \textit{Pedagogical VLA Framework}, a framework that applies pedagogical alignment to lightweight VLA models through four components: text healing to restore language generation capabilities, large language model (LLM) distillation to transfer pedagogical knowledge, safety training for educational environments, and pedagogical evaluation adjusted to science education contexts. We evaluate \textit{Pedagogical VLA Framework} across five science demonstrations spanning physics, chemistry, biology, and earth science, using an evaluation framework developed in collaboration with science education experts. Our evaluation assesses both task performance (success rate, protocol compliance, efficiency, safety) and pedagogical quality through teacher surveys and LLM-as-Judge assessment. We additionally provide qualitative analysis of generated texts. Experimental results demonstrate that \textit{Pedagogical VLA Framework} achieves comparable task performance to baseline models while producing contextually appropriate educational explanations.

\end{abstract}

\begin{figure}
    \centering
    \includegraphics[width=0.9\linewidth]{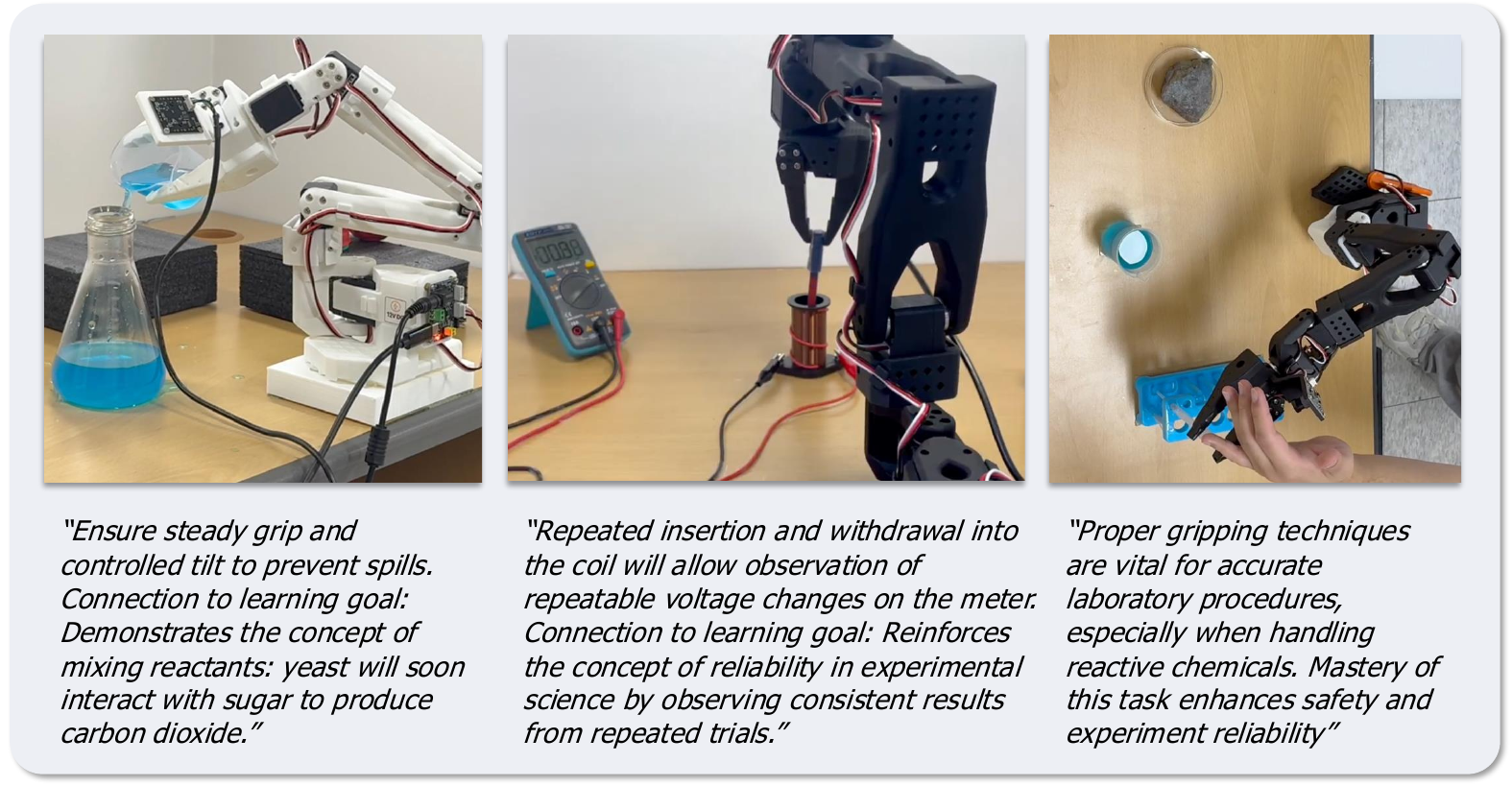}
    \caption{Robot science demonstrations with pedagogically-aligned explanations connecting actions to learning objectives (\textit{left}, \textit{center}), and safety-aware behavior enabling immediate halt upon detecting human presence in the workspace (\textit{right}).}
    \label{fig:first_page}
\end{figure}

\section{Introduction}

\begin{figure*}[t]
\centering
\includegraphics[width=\textwidth]{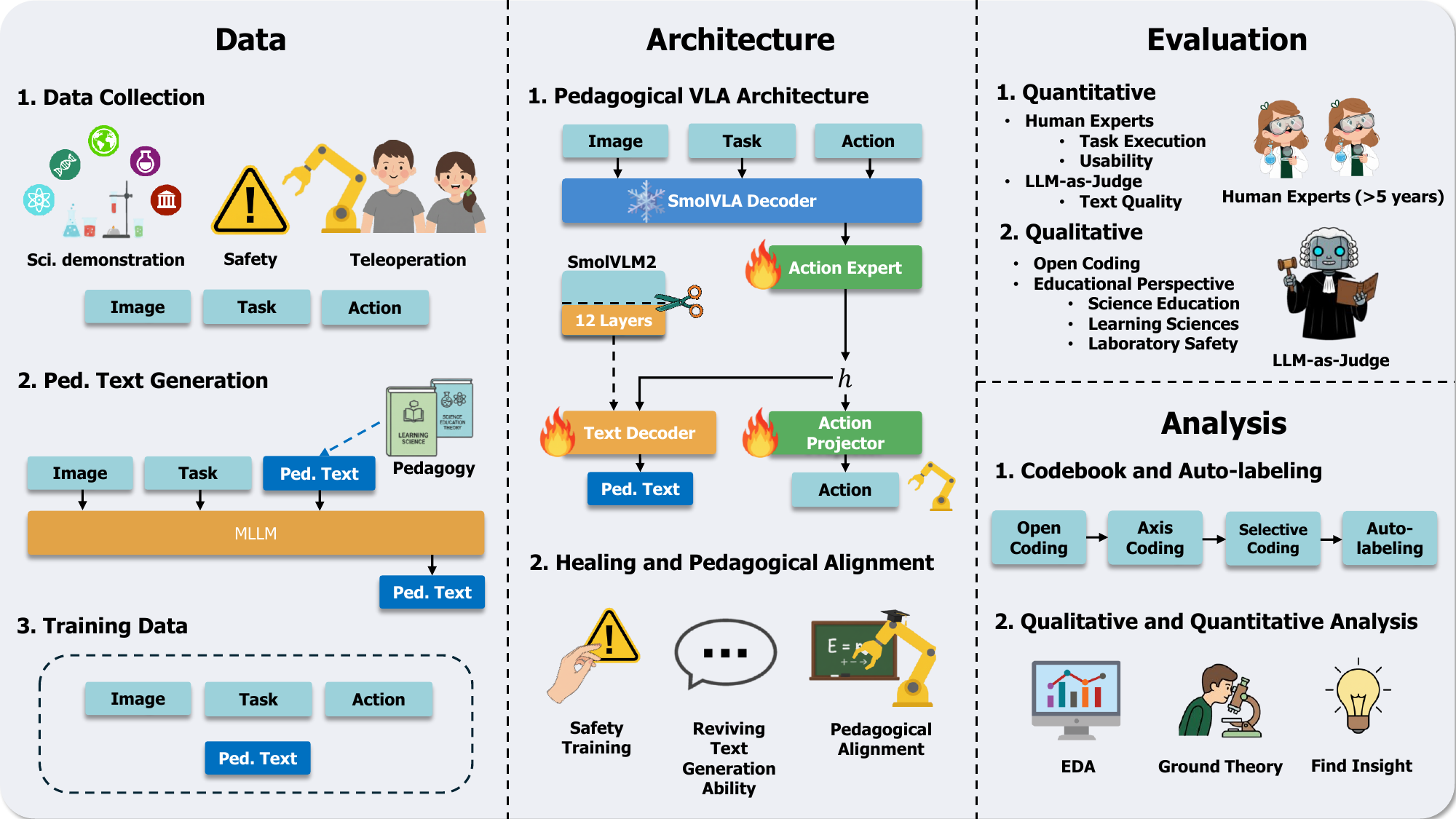}
\caption{Overview of the Pedagogical VLA Framework. The framework integrates data collection via teleoperation with LLM-generated pedagogical text, extends SmolVLA with a text decoder trained for safety awareness and pedagogical alignment, and evaluates both task execution and text quality through human experts and LLM-as-Judge.}
\label{fig:main}
\end{figure*}

Science demonstrations, where teachers perform experiments while students observe, play a crucial role in STEM education by enabling students to witness scientific phenomena closely and connect theoretical concepts to real-world applications \cite{hofstein2004laboratory}. However, conducting these demonstrations poses challenges for teachers, including preparation time, safety concerns when handling hazardous materials, and the difficulty of maintaining consistency across multiple class sessions \cite{singer2006america}.

Robotics has been applied to education primarily through social robots serving as learning companions and rule-based systems for programming education \cite{belpaeme2018social,spolaor2025socialrobots}. These approaches have demonstrated positive effects on student engagement, but they rely on pre-programmed behaviors and lack flexibility to perform complex physical tasks such as science demonstrations.

Recent advances in Vision-Language-Action (VLA) models offer new possibilities for educational robotics \cite{brohan2023rt2,kim2024openvla}. VLA models integrate visual perception, language understanding, and action generation, enabling robots to perform complex manipulation tasks starting from natural language instructions. However, VLA research have focused exclusively on general manipulation tasks \cite{vlasurvey2025}, with almost no consideration of educational contexts. To address this gap, we apply VLA models to real robots in education, selecting science demonstrations as the target domain due to their need for precise manipulation, safety awareness, and pedagogical explanation.

We identifed four critical gaps in applying VLA models to education: (1) state-of-the-art VLA models require substantial computing resources impractical for typical classrooms, necessitating lightweight alternatives; (2) most VLA models sacrifice language generation for efficiency, but educational contexts require models that explain their actions for interpretability; and (3) existing VLA models lack safety mechanisms for settings where students may enter the workspace, and (4) evaluation frameworks neglect pedagogical quality.

To address these gaps, we propose \textit{Pedagogical VLA Framework}\footnote{\url{https://anonymous.4open.science/r/pedagogical_vla_submission-517D/README.md}}, which applies pedagogical alignment to lightweight VLA models through four components: \textit{text healing} to restore language generation capabilities, \textit{Large Language Model (LLM) distillation} to transfer pedagogical knowledge from LLMs, \textit{safety training} to halt operation when students enter the workspace, and \textit{pedagogical evaluation} for multi-dimensional assessment of both task performance and educational quality. We implement this framework by extending SmolVLA \cite{smolvla2024} with a text decoder module that generates pedagogically-aligned explanations during task execution.

\subsection{Contributions}

\begin{itemize}
    \item We propose \textit{Pedagogical VLA Framework}, which applies VLA models to educational contexts based on five design principles derived from educational robotics and pedagogical alignment research.

    \item We introduce text healing and pedagogical data distillation to achieve interpretability, transparency, and pedagogical alignment in lightweight VLA models.

    \item We design a multi-layered evaluation framework that assesses both robot task performance and educational usability through quantitative metrics, teacher surveys, and LLM-as-Judge.
\end{itemize}

\section{Pedagogical VLA Framework}

\subsection{Overview}

Drawing from educational robotics \cite{belpaeme2018social,spolaor2025socialrobots} and pedagogical alignment research \cite{lee2025pedagogy}, we identify five design principles for applying VLA models to education: lightweight-first, interpretability through language, safety-by-design, pedagogy-aligned, and multi-dimensional evaluation (Table~\ref{tab:principles}).

\begin{table}[h!]
\caption{Design principles for pedagogical VLA derived from educational robotics and pedagogical alignment research.}
\centering
\small
\begin{tabular*}{\columnwidth}{@{\extracolsep{\fill}}ll@{}}
\toprule
Principle & Description \\
\midrule
Lightweight-First & Resource-constrained deployment \\
Interpretability & Natural language explanations \\
Safety-by-Design & Integrated safety mechanisms \\
Pedagogy-Alignment & Connection to learning objectives \\
Multi-dimensional & Task and pedagogical evaluation \\
\bottomrule
\end{tabular*}
\label{tab:principles}
\end{table}

Based on these principles, \textit{Pedagogical VLA Framework} can be examined from three dimensions: Data, Architecture, and Evaluation (See figure \ref{fig:main}). The Data dimension addresses how to collect and annotate training data with pedagogical content through LLM distillation and safety intervention scenarios. The Architecture dimension describes how to restore language generation capabilities to lightweight VLA models through text healing. The Evaluation dimension defines how to assess both task performance and educational usability through quantitative metrics and teacher surveys. Moreover, we analyze generated text with exploratory data analsys (EDA) and ground theory \cite{strauss1998basics}

\subsection{Data}

\subsubsection{LLM Distillation for Pedagogical Text}

Educational explanations require domain knowledge about science concepts, pedagogical strategies, and curriculum alignment that cannot be learned from action demonstrations alone. We address this through LLM distillation, transferring pedagogical knowledge from LLMs to our lightweight model.

SmolVLA employs action chunking, predicting 50 actions per inference. We align text generation with this chunking by generating one pedagogical annotation per action chunk using GPT-4o. The annotations follow a structured format:

\begin{tcolorbox}[colback=lightblue, colframe=black, boxrule=0.5pt, arc=2pt, breakable]
\small
\texttt{[Stage]} Current demonstration phase\\
\texttt{[Action]} Robot's current action description\\
\texttt{[Safety Status]} Normal / Stop - Human detected\\[0.5em]
\textbf{Learning Focus}: Key concept being demonstrated\\
\textbf{Connection to learning goal}: How this step relates to learning objectives\\
\textbf{Next}: Upcoming action or instruction
\end{tcolorbox}

We construct few-shot prompts with exemplars covering idle states, active manipulation, and safety intervention scenarios. This ensures the distilled model learns to generate appropriate responses across diverse situations.

\subsubsection{Safety Intervention Data}

Educational environments present unique safety challenges as students may unexpectedly reach into the robot's workspace \cite{robinsafety2020,hrisafety2021}. Unlike industrial settings with physical barriers, classroom deployments must handle such situations gracefully.

We collect episodes where human hands enter the workspace during experiments. The robot outputs zero-velocity commands and generates safety-focused text annotations. This data teaches the model to detect human presence, halt immediately, explain the intervention, and wait for clearance before resuming.

\subsection{Architecture}

\subsubsection{Input Representation}

The model receives three types of inputs at each timestep $t$. Visual observations $I_t^{\text{wrist}}, I_t^{\text{top}} \in \mathbb{R}^{H \times W \times 3}$ are RGB images from wrist-mounted and top-view cameras ($H=480$, $W=640$). These provide complementary perspectives where the wrist camera captures fine manipulation details while the top-view camera provides global scene context. Proprioceptive state $s_t \in [-100, 100]^6$ is a 6-dimensional vector representing normalized joint positions. Task instruction $\ell = (w_1, w_2, \ldots, w_L)$ is a token sequence describing the experimental task, e.g., \textit{``Hold stick, tap powder, hold over blue 3s, return to cup. Stop if hand detected.''}

\subsubsection{Base Model}

We build upon SmolVLA \cite{smolvla2024}, a 450M parameter lightweight VLA derived from the SmolVLM vision-language model. SmolVLA inherits SmolVLM's vision encoder and text model layers but removes the language model head to prioritize action prediction efficiency. The architecture processes inputs as follows.

The vision encoder $f_{\text{vision}}$ (SigLIP, frozen) extracts visual features:
\begin{equation}
z_t^v = f_{\text{vision}}([I_t^{\text{wrist}}; I_t^{\text{top}}])
\end{equation}
The text model layers $f_{\text{backbone}}$ (frozen) fuse visual features, proprioceptive state, and language instruction into a multimodal representation:
\begin{equation}
h_t = f_{\text{backbone}}(z_t^v, s_t, z^\ell)
\end{equation}
The action expert $f_{\text{action}}$ (trainable) predicts action chunks using flow matching:
\begin{equation}
a_t, h_t^{\text{expert}} = f_{\text{action}}(h_t)
\end{equation}
where $a_t \in \mathbb{R}^{C \times D}$ with chunk size $C=50$ and action dimension $D=6$, and $h_t^{\text{expert}}$ denotes the expert's hidden states.

\subsubsection{Text Healing}

While SmolVLA's removal of the language model head improves efficiency, it eliminates interpretability through language output. Text healing restores this capability by reconstructing the text generation pathway using three components.

First, we extract context from the action expert's hidden states through mean pooling:
\begin{equation}
c_t = \text{MeanPool}(h_t^{\text{expert}})
\end{equation}
Second, a projection layer transforms the expert dimension to the language model dimension:
\begin{equation}
c_t^{\text{proj}} = W_{\text{proj}} \cdot c_t
\end{equation}
where $W_{\text{proj}} \in \mathbb{R}^{d_{\text{hidden}} \times d_{\text{expert}}}$ is newly trained.

Third, we restore text generation using components initialized from SmolVLM: a text decoder $f_{\text{dec}}$ (12 transformer layers, trainable) and a language model head $W_{\text{lm}}$ (pretrained weights, trainable):
\begin{equation}
h_i^{\text{dec}} = f_{\text{dec}}(c_t^{\text{proj}}, y_{<i})
\end{equation}
\begin{equation}
P(y_i | y_{<i}, c_t) = \text{softmax}(W_{\text{lm}} \cdot h_i^{\text{dec}})
\end{equation}
The generated sequence $y = (y_1, \ldots, y_T)$ describes the demonstration stage, current action, safety status, learning focus, and upcoming steps. The text decoder is trained on pedagogical annotations from LLM distillation (Section 2.2.1) and safety intervention data (Section 2.2.2), enabling the model to produce both educationally meaningful explanations and appropriate safety responses

\subsubsection{Training Objective}

The model is trained with a combined loss function:

\begin{equation}
\mathcal{L}_{\text{total}} = \mathcal{L}_{\text{action}} + \lambda \mathcal{L}_{\text{text}}
\end{equation}

where

\begin{equation}
\mathcal{L}_{\text{action}} = \text{MSE}(a_t, \hat{a}_t)
\end{equation}

is the mean squared error between predicted and ground-truth actions, and

\begin{equation}
\mathcal{L}_{\text{text}} = \text{CE}(y, \hat{y})
\end{equation}

is the cross-entropy loss for text generation. The weight $\lambda \in [0.01, 1.0]$ balances action accuracy and text quality; we set $\lambda = 0.1$ based on preliminary experiments.

\subsection{Evaluation}

\subsubsection{Quantitative Metrics}

We adopt evaluation metrics from recent laboratory automation benchmarks \cite{autobio2024,chemistry3d2025,labbench2025}. Table~\ref{tab:quant_metrics} summarizes the metrics organized into three categories: task execution, text quality, and usability. For each task, we run 10 episodes and average the scores. Task execution metrics are rated by two human experts who evaluate independently and then resolve disagreements through discussion. Text quality is assessed using LLM-as-Judge with GPT-4o. Usability is evaluated by 2 science education experts using a 20-item Likert scale questionnaire (Appendix~\ref{app:survey}). Detailed evaluation criteria are provided in Appendix~\ref{app:criteria} and the rubric in Appendix~\ref{app:rubric}.

\begin{table}[H]
\caption{Quantitative evaluation metrics adapted from laboratory automation benchmarks to assess both task performance and educational usability.}
\centering
\scriptsize
\begin{tabular}{p{1.1cm}p{1.1cm}p{3.1cm}cc}
\toprule
Group & Category & Metrics & Eval. & Scale \\
\midrule
Task\newline Execution & Task\newline Success & Grip, placement, task-specific & Human & Acc. \\
& Protocol & Step order, condition compliance & Human & Acc. \\
& Efficiency & Unnecessary action rate & Human & Acc. \\
& Safety & Collision, drop, jerk, human detection & Human & Acc. \\
\midrule
Text\newline Quality & -- & Relevance, pedagogical value, safety comm., fluency & LLM & 1-5 \\
\midrule
Usability & -- & Effectiveness, efficiency, safety, sustainability, enjoyment & Human & 1-5 \\
\bottomrule
\end{tabular}
\label{tab:quant_metrics}
\end{table}

\subsubsection{Qualitative Assessment}

Two science education experts record narrative observations during the quantitative evaluation process. These observations capture contextual factors, failure modes, and pedagogical insights that numerical metrics cannot fully represent. We analyze these expert notes using thematic synthesis to identify patterns across demonstrations and models.

\section{Experiments}

\subsection{Experimental Settings}

\subsubsection{Demonstration Tasks}

Two science education experts within our research team selected five demonstration tasks most suitable for robot-assisted demonstrations in K-12 science education. The selection criteria included curriculum relevance, procedural clarity, and safety considerations. Table~\ref{tab:tasks} summarizes the selected tasks spanning major STEM disciplines.

\begin{table}[h!]
\caption{Overview of demonstration tasks selected based on curriculum relevance, procedural clarity, and safety considerations.}
\centering
\tiny
\begin{tabular*}{\columnwidth}{@{\extracolsep{\fill}}llp{3.2cm}@{}}
\toprule
Domain & Demonstration & Description \\
\midrule
Physics & Electromagnetic Induction & Magnet insertion/withdrawal through coil \\
Chemistry & Flame Test & Wire coating, heating to observe emission colors \\
Biology & Yeast Fermentation & Dispensing, sealing flasks to observe CO$_2$ \\
Earth Sci. & Rock Classification & Acid application to identify carbonate \\
Lab Support & Agar Plate Prep. & Pouring agar into petri dishes \\
\bottomrule
\end{tabular*}
\label{tab:tasks}
\end{table}

\subsubsection{Data Collection}

We used the SO-101 robot arm, a low-cost 6-DOF manipulator with Feetech STS3215 servo motors (Appendix~\ref{app:config}). We collected dual-camera video (wrist and top view) and action trajectories through teleoperation for each task. For each demonstration, we collected both normal execution episodes and safety intervention scenarios where a human hand enters the workspace. Text annotations were generated separately: simple action descriptions for Text-SmolVLA baseline, and pedagogical explanations via GPT-4o with task-specific few-shot prompts for Pedagogical VLA. Table~\ref{tab:dataset} summarizes the dataset statistics.

\begin{table}[h!]
\caption{Dataset statistics showing episode distribution across safety intervention and normal conditions. All data collected at 30 fps with dual-camera setup.}
\centering
\small
\begin{tabular*}{\columnwidth}{@{\extracolsep{\fill}}lrrr@{}}
\toprule
Demonstration & Safety Intervention & Normal & Total \\
\midrule
EM Induction & 20 & 60 & 80 \\
Flame Test & 20 & 60 & 80 \\
Yeast Fermentation & 22 & 100 & 122 \\
Rock Classification & 30 & 100 & 130 \\
Agar Plate Prep. & 30 & 130 & 160 \\
\bottomrule
\end{tabular*}
\label{tab:dataset}
\end{table}

\subsubsection{Baseline Models}

We compare \textit{Pedagogical VLA Framework} against three baselines. ACT \cite{zhao2023act} is the Action Chunking Transformer, a state-of-the-art imitation learning method without language capabilities. SmolVLA \cite{smolvla2024} is a lightweight VLA model that conditions on language instructions but does not generate text outputs. Text-SmolVLA is SmolVLA augmented with our text decoder but trained to generate simple action descriptions (e.g., \textit{``Picking up the beaker''}) rather than pedagogical explanations, serving as an ablation to isolate the effect of LLM distillation.

\subsubsection{Training Details}

All models are trained for 100,000 steps with batch size 32. The text generation model for distillation is GPT-4o. The text loss weight $\lambda$ is set to 0.1, maximum text length to 128 tokens, and text decoder depth to 12 layers.

\begin{table*}[t]
\caption{Quantitative evaluation results. \colorbox{gray!15}{Gray} indicates non-text models; \colorbox{cyan!10}{Blue} indicates text-generating models. Task execution metrics are in \% (higher is better), except Eff.$\downarrow$ (lower is better). Det. is Human Detection Stop Rate; Manip. is Manipulation Safety (100\% $-$ avg. of collision, drop, jerk rates). Text Quality and Usability are on 1-5 scale. Results averaged over 5 experiments and 10 episodes per task.}
\centering
\resizebox{\textwidth}{!}{%
\begin{tabular}{lccccccccccccccc}
\toprule
& \multicolumn{6}{c}{Task Execution (\%)} & \multicolumn{4}{c}{Text Quality (1-5)} & \multicolumn{5}{c}{Usability (1-5)} \\
\cmidrule(lr){2-7} \cmidrule(lr){8-11} \cmidrule(lr){12-16}
& Task & \multicolumn{2}{c}{Protocol} & & \multicolumn{2}{c}{Safety} & & & & & & & & & \\
\cmidrule(lr){3-4} \cmidrule(lr){6-7}
Model & Succ. & Order & Cond. & Eff.$\downarrow$ & Det. & Manip. & Rel. & Ped. & Safe & Flu. & Eff. & Effi. & Safe & Sust. & Enj. \\
\midrule
\rowcolor{gray!15} \multicolumn{16}{l}{\textit{Non-text Models}} \\
ACT & \textbf{78} & \textbf{80} & \textbf{64} & \textbf{26} & 20 & 80 & -- & -- & -- & -- & \textbf{3.7} & \textbf{4.2} & \textbf{3.2} & \textbf{3.8} & \textbf{4.9} \\
SmolVLA & 49 & 42 & 31 & 52 & \textbf{78} & 76 & -- & -- & -- & -- & 2.5 & 2.8 & \textbf{3.2} & 2.8 & 4.1 \\
\midrule
\rowcolor{cyan!10} \multicolumn{16}{l}{\textit{Text-generating Models}} \\
Text-SmolVLA & \textbf{38} & \textbf{28} & 20 & \textbf{40} & \textbf{78} & 85 & 2.3 & 1.3 & 1.0 & \textbf{3.9} & \textbf{2.8} & \textbf{3.3} & \textbf{3.3} & \textbf{3.1} & 3.5 \\
Pedagogical VLA (Ours) & 29 & 22 & \textbf{24} & 48 & 60 & \textbf{90} & \textbf{3.7} & \textbf{3.5} & \textbf{2.7} & 2.6 & 2.6 & 3.1 & 3.2 & \textbf{3.1} & \textbf{3.8} \\
\bottomrule
\end{tabular}%
}
\label{tab:quant_results}
\end{table*}

\subsection{Experiment Results}

\subsubsection{Quantitative Evaluation}

Table~\ref{tab:quant_results} presents quantitative evaluation results across three categories. Task execution metrics (Task Success, Protocol, Efficiency, Safety) are rated by human experts as accuracy (\%). Text quality is assessed by LLM-as-Judge on a 1-5 scale. Usability is evaluated by 2 science teachers (each with $>$5 years experience) using a 5-point Likert scale.

For task execution, ACT achieves the highest scores across Task Success (78\%), Protocol Order (80\%), and Condition Compliance (64\%), demonstrating the effectiveness of action-focused imitation learning without text generation overhead. SmolVLA shows moderate performance (49\% task success), while Text-SmolVLA and Pedagogical VLA exhibit lower task success (38\% and 29\%) due to the additional text generation objective competing for model capacity. Notably, VLA models with text output achieve higher Human Detection Stop rates (60--78\%) compared to ACT (20\%), suggesting that language conditioning improves safety response learning. Pedagogical VLA achieves the highest Manipulation Safety score (90\%), computed as 100\% minus the average of collision, drop, and jerk event rates.

For text quality evaluation, we used GPT-4o-mini as an LLM-as-Judge to assess 1,000 generated utterances (50 samples per condition) across four metrics. Pedagogical VLA significantly outperformed Text-SmolVLA in Relevance (3.7 vs 2.3), Pedagogical Value (3.5 vs 1.3), and Safety Communication (2.7 vs 1.0), with all differences statistically significant ($p$ < .001). Text-SmolVLA achieved higher Fluency scores (3.9 vs 2.6) because its simple action descriptions contain fewer grammatical errors than the longer pedagogical explanations. These results demonstrate that LLM distillation successfully transfers educational scaffolding capabilities to the lightweight model. Detailed results by experiment and condition are provided in Appendix~\ref{app:text_quality}.

\subsubsection{Qualitative Evaluation}

Two experts within our research team provided narrative observations. A recurring theme was safety detection: evaluators noted ``ACT lacks safety response,'' while VLA models' detection required close proximity, with one noting ``if the robot only stops when hands are very close, it seems meaningless.''

Evaluators identified a monitoring burden when success rates were low: ``If the robot arm frequently fails, I would have to continuously monitor it. Then I cannot focus on students.'' This was amplified for liquid-handling tasks, where ``solution spillage frequency is very high.''

Despite challenges, evaluators recognized unique value. For hazardous experiments, ``significant time is spent on safety instructions; this time can be reduced'' by robots. They appreciated fine motor capabilities: ``The robot performed well at fine gap adjustments required for dropper use,'' but cautioned that reliability must improve before classroom deployment.

\begin{figure*}[t]
\centering
\begin{minipage}{0.32\textwidth}
    \centering
    \includegraphics[width=\textwidth]{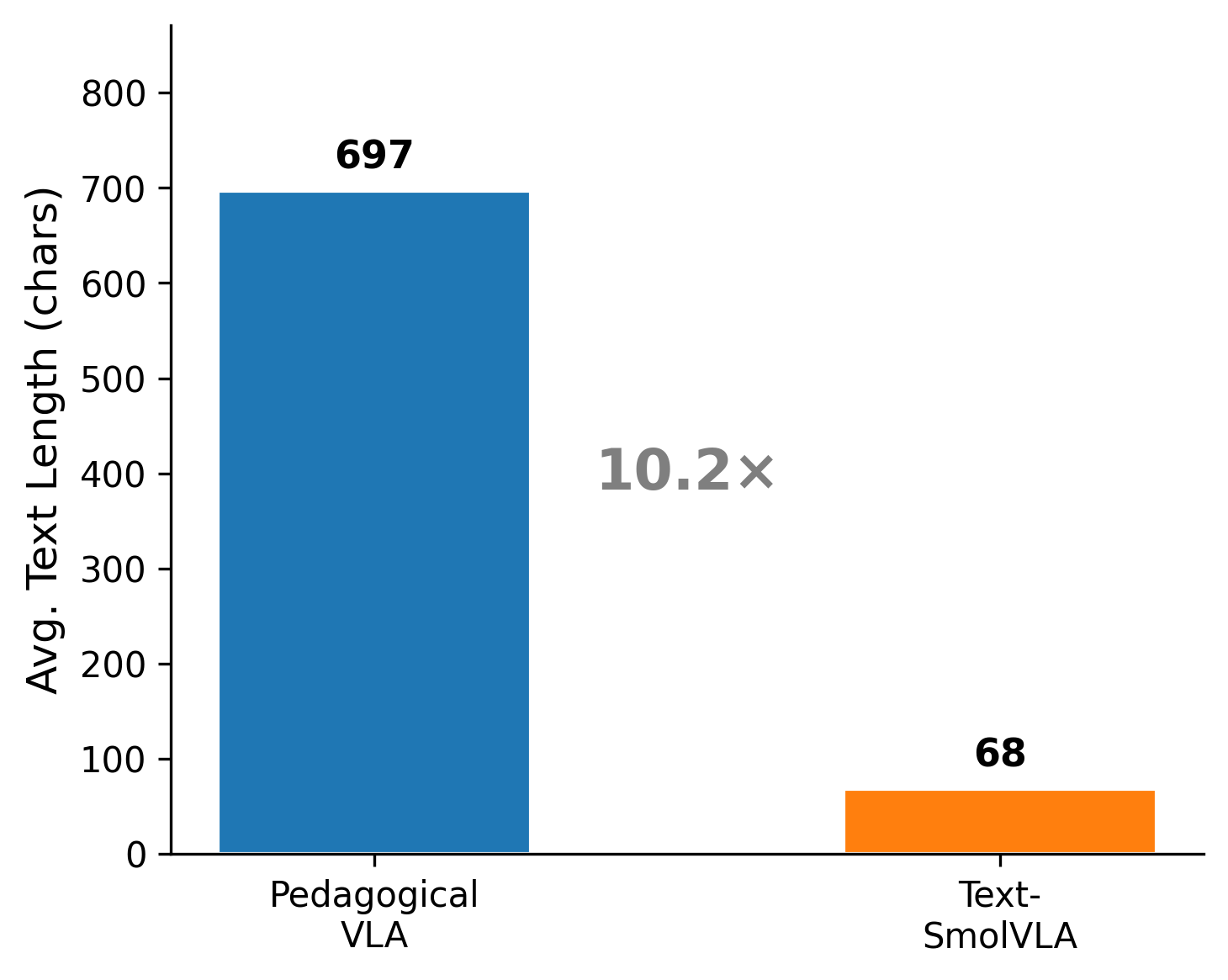}
\end{minipage}
\hfill
\begin{minipage}{0.32\textwidth}
    \centering
    \includegraphics[width=\textwidth]{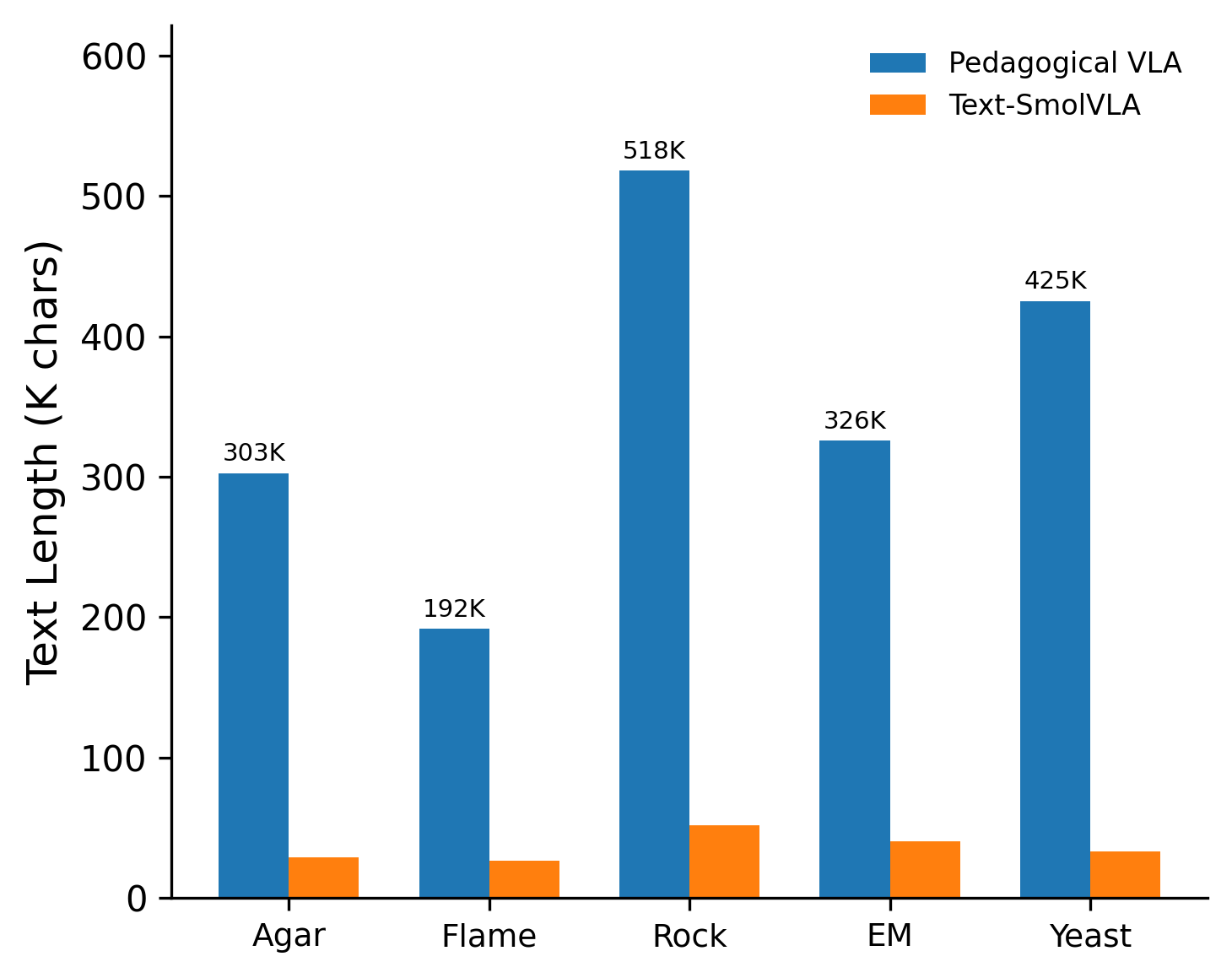}
\end{minipage}
\hfill
\begin{minipage}{0.32\textwidth}
    \centering
    \includegraphics[width=\textwidth]{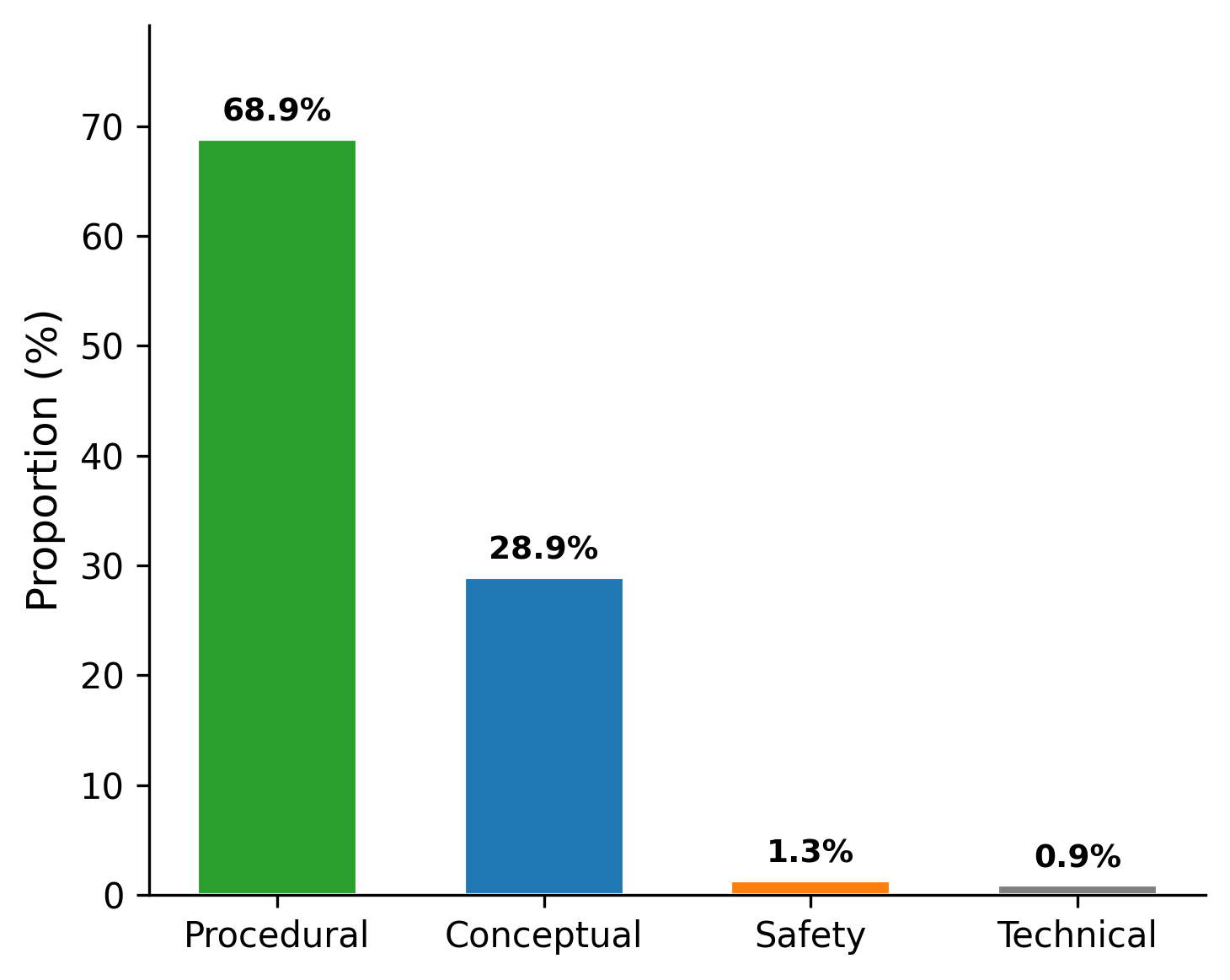}
\end{minipage}
\caption{Descriptive statistics of generated text. The \textit{left} panel shows average text length by model, where Pedagogical VLA generates 10.2$\times$ longer utterances (697 vs 68 characters). The \textit{center} panel compares text volume by experiment between Pedagogical VLA (blue) and Text-SmolVLA (orange). The \textit{right} panel shows Pedagogical VLA category distribution with Procedural Support (68.9\%) and Conceptual Support (28.9\%) as dominant categories.}
\label{fig:descriptive}
\end{figure*}

\section{Generated Text Analysis}

\begin{figure*}[t]
\centering
\begin{minipage}{0.32\textwidth}
    \centering
    \includegraphics[width=\textwidth]{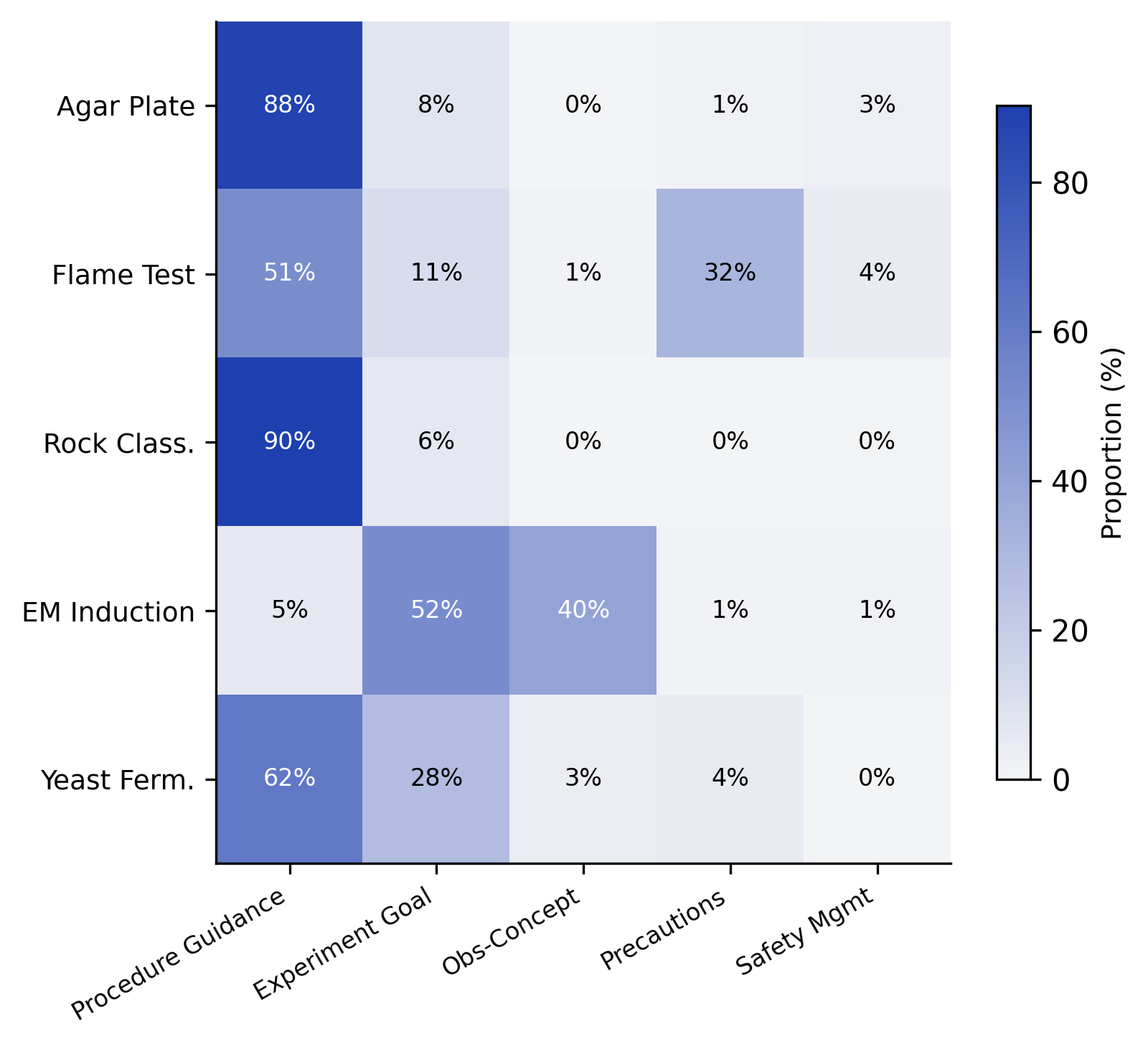}
\end{minipage}
\hfill
\begin{minipage}{0.32\textwidth}
    \centering
    \includegraphics[width=\textwidth]{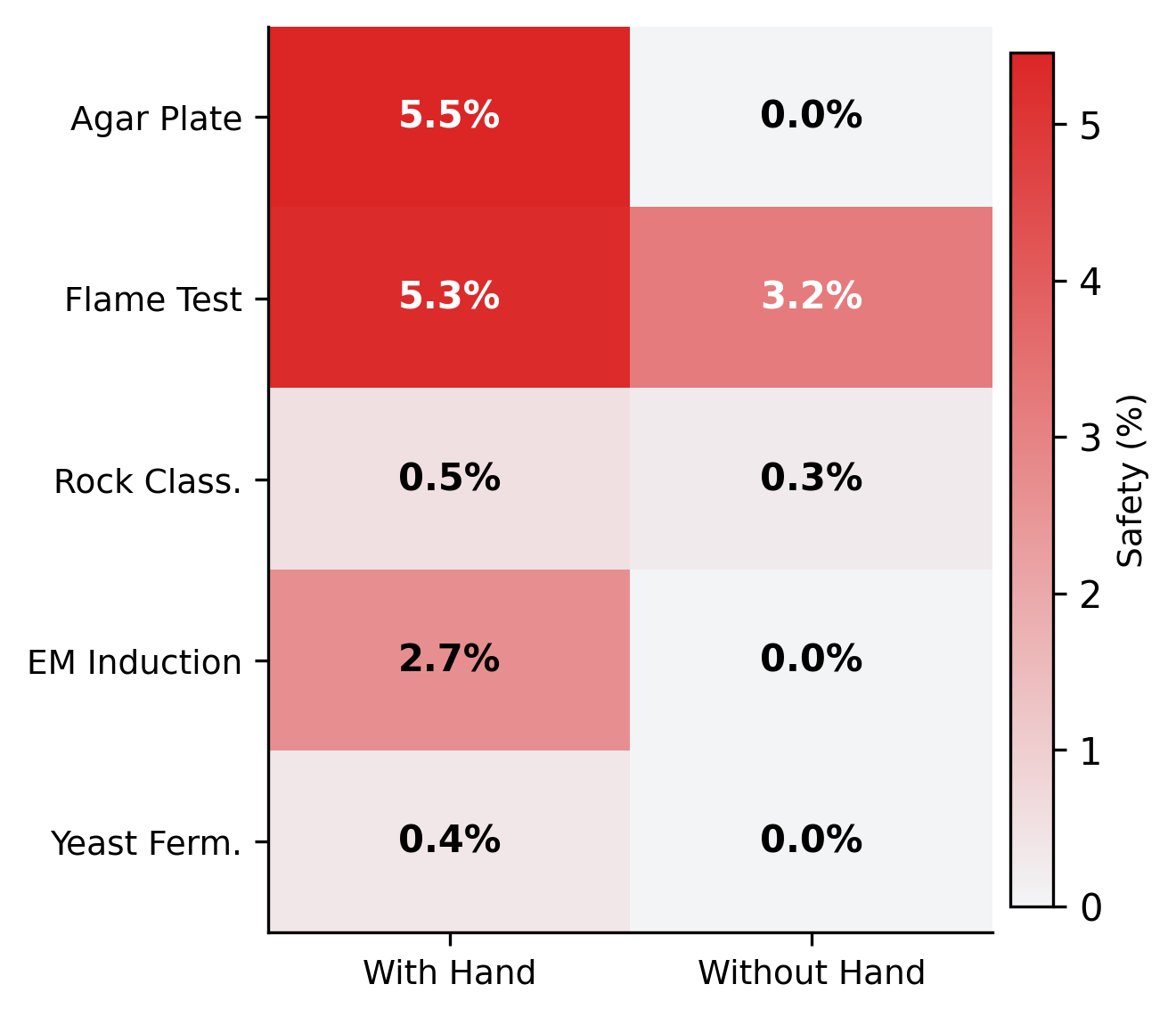}
\end{minipage}
\hfill
\begin{minipage}{0.32\textwidth}
    \centering
    \includegraphics[width=\textwidth]{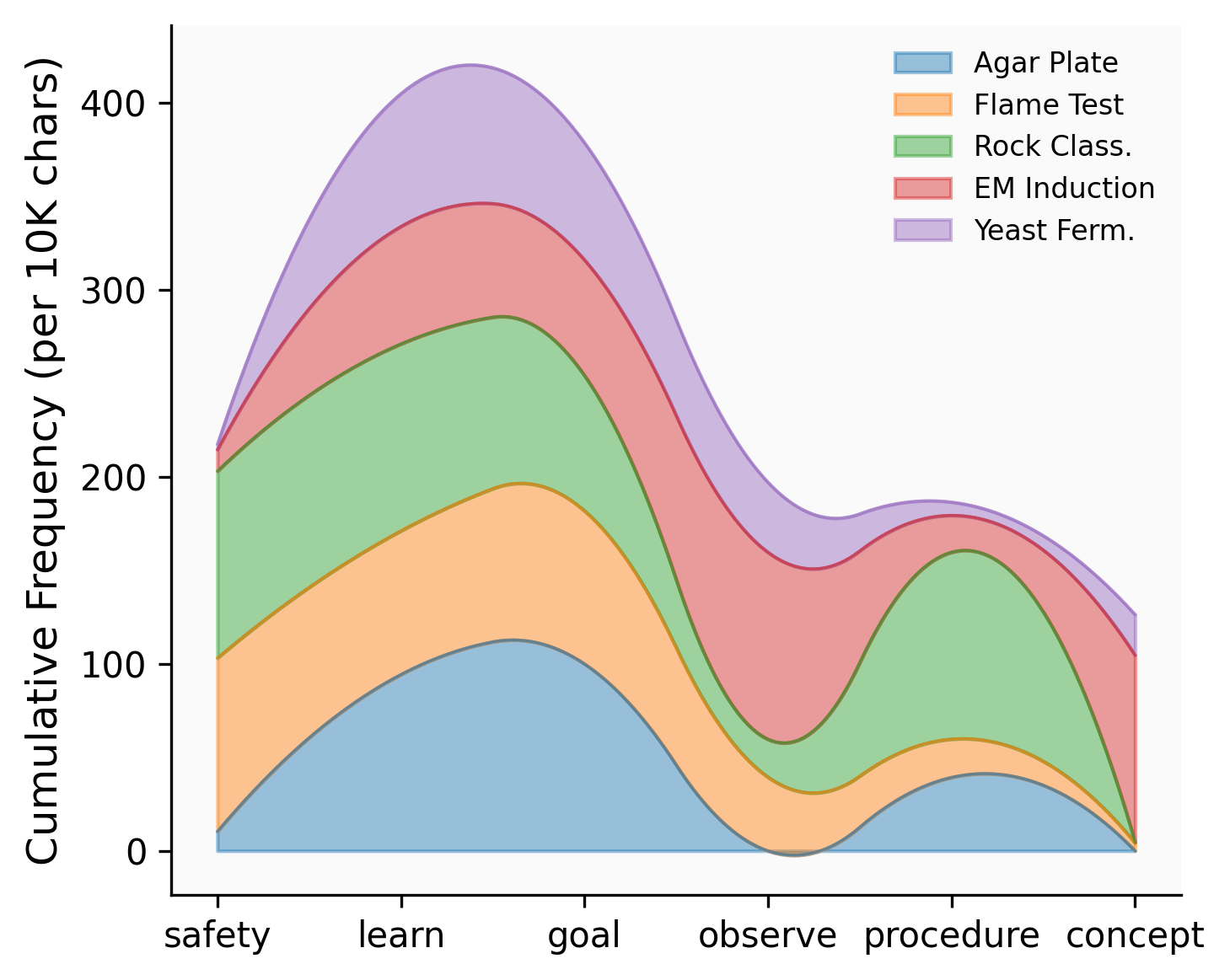}
\end{minipage}
\caption{Main analysis results. The \textit{left} panel shows experiment-specific subcategory patterns where EM Induction emphasizes Experiment Goal (52\%) and Observation-Concept (40\%), while Rock Classification emphasizes Procedure Guidance (90\%). The \textit{center} panel presents Safety Management by experiment and hand condition, revealing inherent safety awareness in Flame Test (3.2\% without hand) versus reactive safety in other experiments (0\% without hand). The \textit{right} panel displays keyword frequency patterns by experiment showing distinct pedagogical emphases across tasks.}
\label{fig:analysis}
\end{figure*}

\subsection{Codebook Development}

We developed a codebook for analyzing pedagogical text outputs following grounded theory methodology \cite{strauss1998basics}. Initial codes were drafted based on established science education frameworks: scaffolding theory \cite{vandepol2010scaffolding}, the BSCS 5E instructional model \cite{bybee2006bscs}, inquiry-based science education \cite{vanuum2021inquiry,petersen2022scaffolding}, and laboratory instruction research \cite{hofstein2004laboratory,millar2004practical,singer2006america}. Two researchers then independently reviewed model-generated text outputs for each episode, iteratively refining codes through discussion until reaching consensus.

Table~\ref{tab:codebook} presents the final codebook with six categories spanning conceptual, procedural, inquiry, affective, safety, and technical dimensions.

\begin{table}[H]
\caption{Codebook for pedagogical text analysis developed through grounded theory methodology based on established science education frameworks.}
\centering
\tiny
\begin{tabular*}{\columnwidth}{@{\extracolsep{\fill}}llp{3.8cm}@{}}
\toprule
Category & Code & Definition \\
\midrule
\multirow{2}{*}{Conceptual} & EXP-GOAL & States learning objective or what will be verified \\
& OBS-CONCEPT & Connects observations to scientific concepts \\
\midrule
\multirow{3}{*}{Procedural} & PROC-GUIDE & Guidance on experimental procedures \\
& EQUIP-OP & Instructions for equipment operation \\
& PRECAUTION & Emphasis on accuracy/reliability precautions \\
\midrule
\multirow{4}{*}{Inquiry} & HYPOTHESIS & Predictions about experiment results \\
& VAR-CTRL & Explains variable control rationale \\
& RESULT-OBS & Guidance for observing results \\
& RESULT-INT & Interprets results or guides conclusions \\
\midrule
Affective & AFF-SUPPORT & Addresses learner anxiety or motivation \\
\midrule
Safety & SAFETY-MGT & Emphasizes safety to prevent accidents \\
\midrule
Technical & TECH-CAP & Highlights robotic system capabilities \\
\bottomrule
\end{tabular*}
\label{tab:codebook}
\end{table}

\subsection{Quantitative Analysis}

We classified all generated text using GPT-4o-mini, measuring character counts to account for length differences between models (Figure~\ref{fig:descriptive}). Pedagogical VLA generated 90.8\% of total characters, with 99.1\% classified as educational scaffolding. Text-SmolVLA generated 9.2\%, with 99.9\% classified as Technical Execution, demonstrating effective pedagogical knowledge transfer through LLM distillation.

The \textit{left} panel of Figure~\ref{fig:analysis} shows how scaffolding subcategory patterns vary across experiments. EM Induction emphasizes Experiment Goal (52\%) and Observation-Concept Connection (40\%), reflecting the need to explain electromagnetic principles. Rock Classification shows 90\% Procedure Guidance, emphasizing step-by-step instructions for acid application and observation. These results suggest that Pedagogical VLA adapts its scaffolding strategy to match the pedagogical requirements of each experiment.

The \textit{center} panel of Figure~\ref{fig:analysis} reveals safety awareness patterns across experiments and hand conditions. When a human hand entered the robot workspace, Safety Management comprised 2.29\% of generated text compared to 0.39\% without hand presence (Odds Ratio = 5.49, $p$ < .001). Most experiments show reactive safety behavior, generating safety utterances only when a hand enters the workspace. In contrast, Flame Test exhibits inherent safety awareness, producing safety content even without hand intervention (3.2\% without hand, 5.3\% with hand), suggesting the model acquired prior knowledge about experiment-specific hazards through distillation.

The \textit{right} panel of Figure~\ref{fig:analysis} shows keyword frequency patterns across experiments. EM Induction shows high frequency of learning-related keywords, reflecting its conceptual focus. Flame Test emphasizes safety-related terms. These patterns demonstrate that Pedagogical VLA generates contextually appropriate vocabulary aligned with each experiment's pedagogical goals.

\subsection{Qualitative Analysis}

Manual coding of representative outputs reveals distinct scaffolding patterns. Pedagogical VLA generates conceptual support linking observations to scientific principles \cite{vandepol2010scaffolding}: e.g., \textit{``Helps visualize how consistent motions affect the induced voltage, reinforcing electromagnetic induction.''} It also provides safety scaffolding \cite{hofstein2004laboratory}: e.g., \textit{``Maintain a safe distance from automated equipment.''} In contrast, Text-SmolVLA produces only technical execution descriptions across all experiments: e.g., \textit{``The robot is pouring blue liquid from a beaker into a flask''} and \textit{``The robot is dispensing a clear liquid from a pipette onto a rock.''}

This pattern aligns with scaffolding theory's distinction between cognitive and procedural support \cite{vandepol2010scaffolding}. Pedagogical VLA provides cognitive scaffolding through concept connections, while Text-SmolVLA's outputs function as mere action narration without educational value. The structured format of Pedagogical VLA outputs, including Learning Focus and Connection to Learning Goal, enables systematic delivery of pedagogical content synchronized with robot actions.

\section{Ablation Study}

\subsection{Text Decoder Depth}

We vary the number of transformer layers in the text decoder module (4, 8, 12 layers) while keeping other hyperparameters fixed. Results in Table~\ref{tab:ablation_depth} show that shallower decoders achieve substantially higher action success rates (80\% for 4 layers vs. 30\% for 12 layers), while text quality peaks at intermediate depth (8 layers). This suggests that excessive decoder capacity may interfere with action prediction while providing diminishing returns for text generation. The 4-layer configuration represents a Pareto-optimal choice when action reliability is prioritized.

\begin{table}[h!]
\caption{Effect of text decoder depth on the trade-off between action success and text quality. Shallow decoders favor action success; intermediate depth optimizes text quality.}
\centering
\small
\begin{tabular}{lcccc}
\toprule
Decoder & Text & Action & Time & Success \\
Layers & Quality ($\uparrow$) & Success ($\uparrow$) & (s) ($\downarrow$) & Time (s) ($\downarrow$) \\
\midrule
4 & \underline{2.41} & \textbf{80} & 35.5 & 37.9 \\
8 & \textbf{2.43} & \underline{40} & \underline{32.5} & \textbf{37.2} \\
12 & 2.32 & 30 & \textbf{25.6} & \underline{37.7} \\
\bottomrule
\end{tabular}
\label{tab:ablation_depth}
\end{table}

\subsection{Text Loss Weight}

We examine the effect of the text loss weight $\lambda$ on the balance between action accuracy and text quality. As shown in Table~\ref{tab:ablation_lambda}, lower $\lambda$ values strongly favor action success (60\% at $\lambda$=0.005 vs. 30\% at $\lambda$=0.05), while higher values improve text quality (2.63 vs. 2.47). This trade-off reflects the competing objectives in the joint loss function: emphasizing text generation diverts model capacity from action prediction. Notably, successful trials with higher $\lambda$ complete faster (38.7s vs. 44.2s), suggesting that better verbal guidance may improve execution efficiency when tasks succeed. The choice of $\lambda$ thus depends on deployment priorities along the Pareto frontier.

\begin{table}[h!]
\caption{Effect of text loss weight $\lambda$ on balancing competing objectives in the joint loss function. Lower $\lambda$ favors action success; higher $\lambda$ improves text quality.}
\centering
\small
\begin{tabular}{lcccc}
\toprule
$\lambda$ & Text & Action & Time & Success \\
 & Quality ($\uparrow$) & Success ($\uparrow$) & (s) ($\downarrow$) & Time (s) ($\downarrow$) \\
\midrule
0.005 & \underline{2.47} & \textbf{60} & 36.9 & 44.2 \\
0.01 & 2.43 & \underline{50} & \underline{35.0} & \underline{40.0} \\
0.05 & \textbf{2.63} & 30 & \textbf{27.9} & \textbf{38.7} \\
\bottomrule
\end{tabular}
\label{tab:ablation_lambda}
\end{table}

\section{Related Work}

Educational robotics research has focused on robots as learning companions or programming platforms \cite{papert1980mindstorms,belpaeme2018social}, while laboratory automation benchmarks \cite{autobio2024,labbench2025} address robotic tasks without educational contexts. However, these approaches have not yet leveraged VLA models, which represent a paradigm shift in robot learning by unifying perception, language understanding, and action generation within single architectures. RT-2 \cite{brohan2023rt2} demonstrated that large vision-language models can be finetuned for robotic control with impressive generalization, OpenVLA \cite{kim2024openvla} introduced an open-source alternative, and SmolVLA \cite{smolvla2024} showed that competitive performance is achievable at smaller scales.

Existing VLA models focus primarily on action generation, often sacrificing language output capabilities to improve efficiency. Our work addresses this limitation through text healing, restoring language generation to lightweight VLA models while adding pedagogical alignment through LLM distillation, bridging the gap between educational robotics and VLA research.

\section{Conclusion}

We presented Pedagogical VLA Framework for science education robotics. The framework restores language generation to lightweight VLA models through text healing, transfers pedagogical knowledge through LLM distillation, and ensures classroom safety through intervention training. Experiments across five science domains demonstrate that Pedagogical VLA achieves comparable task performance while generating contextually appropriate educational explanations with explicit learning objectives and safety guidance. We release our model and dataset to support further research at the intersection of robotics and education.



\bibliographystyle{named}

\clearpage
\appendix

\section{Detailed Evaluation Criteria}
\label{app:criteria}

This appendix provides the complete evaluation criteria for each demonstration task, developed in collaboration with science education experts and laboratory operators.

\subsection{Electromagnetic Induction (Physics)}

This experiment demonstrates Faraday's law through magnet-coil interaction. The robot picks up a bar magnet, inserts it into a solenoid coil connected to a galvanometer, and oscillates it at varying speeds to show how flux change rate affects induced current.

\begin{table}[!htb]
\caption{Evaluation criteria for electromagnetic induction demonstration.}
\centering
\small
\resizebox{\columnwidth}{!}{%
\begin{tabular}{@{}lp{3.5cm}l@{}}
\toprule
Metric & Criterion & Benchmark \\
\midrule
\multicolumn{3}{l}{\textit{Task 1: Magnet Pickup}} \\
Grip Success & Stable magnet grip & $\geq$ 90\% \\
Placement & Position in coil center & $\pm$ 1.5 cm \\
Stability & No drop during transfer & 0\% \\
\midrule
\multicolumn{3}{l}{\textit{Task 2: Oscillation}} \\
Alignment & Magnet centered in coil & $\pm$ 1.5 cm \\
Collision & Contact with coil walls & 0 events \\
Cycle Success & Complete up-down cycle & $\geq$ 85\% \\
Induced Current & Galvanometer response & $\geq$ 5\% \\
\midrule
\multicolumn{3}{l}{\textit{Task 3: Variable Speed}} \\
Cycle Time & Fast oscillation period & $\leq$ 1.5 sec \\
Current Gain & Higher induced signal & $\geq$ 30\% \\
\bottomrule
\end{tabular}%
}
\label{tab:app_em}
\end{table}

\subsection{Flame Test (Chemistry)}

This experiment identifies metal ions through characteristic flame colors. The robot dips a nichrome wire loop into cleaning solution, collects salt samples, holds them in a Bunsen burner flame, and observes emission spectra colors (e.g., sodium: yellow, copper: green).

\begin{table}[!htb]
\caption{Evaluation criteria for flame test demonstration.}
\centering
\small
\resizebox{\columnwidth}{!}{%
\begin{tabular}{@{}lp{3.5cm}l@{}}
\toprule
Metric & Criterion & Benchmark \\
\midrule
\multicolumn{3}{l}{\textit{Task 1: Sample Collection}} \\
Contact Accuracy & Wire tip to sample & $\pm$ 3 mm \\
Sample Adhesion & Single contact attach & $\geq$ 90\% \\
\midrule
\multicolumn{3}{l}{\textit{Task 2: Flame Heating}} \\
Flame Alignment & Wire in flame center & $\pm$ 5 mm \\
Heating Duration & Maintained in flame & 3 $\pm$ 1 sec \\
Color Detection & Observable emission & $\geq$ 30\% \\
\midrule
\multicolumn{3}{l}{\textit{Task 3: Wire Cleaning}} \\
Immersion Depth & Wire submerged & 5-10 mm \\
Cleaning Duration & In cleaning solution & $\geq$ 1.0 sec \\
Residue Removal & Remaining sample & $\leq$ 10\% \\
\bottomrule
\end{tabular}%
}
\label{tab:app_flame}
\end{table}

\subsection{Yeast Fermentation (Biology)}

This experiment demonstrates cellular respiration by comparing fermentation rates with different sugar concentrations. The robot dispenses sugar and yeast into flasks, adds water, seals with balloon-capped lids, and students observe CO$_2$ production through balloon inflation over time.

\begin{table}[!htb]
\caption{Evaluation criteria for yeast fermentation demonstration.}
\centering
\small
\resizebox{\columnwidth}{!}{%
\begin{tabular}{@{}lp{3.5cm}l@{}}
\toprule
Metric & Criterion & Benchmark \\
\midrule
\multicolumn{3}{l}{\textit{Task 1: Sugar Dispensing}} \\
Amount & Flask 1: 1, Flask 2: 2 & $\pm$ 0.1 spoon \\
Placement & Flask opening center & $\pm$ 3 mm \\
Spillage & Material outside flask & 0 g \\
\midrule
\multicolumn{3}{l}{\textit{Task 2: Yeast \& Water Addition}} \\
Yeast Amount & 1 spoon per flask & $\pm$ 0.1 spoon \\
Water Volume & 100 mL per flask & $\pm$ 5 mL \\
Mixing & No layer separation & $\leq$ 10\% \\
\midrule
\multicolumn{3}{l}{\textit{Task 3: Cap Sealing}} \\
Alignment & Cap centered on opening & $\leq$ 2 mm \\
Rotation & Turns to seal & 1.5-2.5 \\
Balloon & No damage or leakage & 0 failures \\
\bottomrule
\end{tabular}%
}
\label{tab:app_yeast}
\end{table}

\subsection{Rock Classification (Earth Science)}

This experiment distinguishes carbonate rocks (limestone, marble) from non-carbonite rocks using acid tests. The robot picks up a dropper, aspirates dilute hydrochloric acid, applies drops to rock samples, and observes fizzing reactions indicating calcium carbonate presence.

\begin{table}[!htb]
\caption{Evaluation criteria for rock classification demonstration.}
\centering
\small
\resizebox{\columnwidth}{!}{%
\begin{tabular}{@{}lp{3.5cm}l@{}}
\toprule
Metric & Criterion & Benchmark \\
\midrule
\multicolumn{3}{l}{\textit{Task 1: Acid Application}} \\
Drop Accuracy & Acid on rock surface & $\pm$ 3 mm \\
Drop Volume & Single drop applied & 0.05 mL \\
Spillage & Acid outside target & 0 drops \\
\midrule
\multicolumn{3}{l}{\textit{Task 2: Observation}} \\
Wait Duration & Observation period & 5 $\pm$ 1 sec \\
Position Hold & Stable viewing angle & $\pm$ 5 deg \\
\midrule
\multicolumn{3}{l}{\textit{Task 3: Classification}} \\
Fizz Detection & CO$_2$ bubble observed & Yes/No \\
Sort Accuracy & Correct bin placement & 100\% \\
\bottomrule
\end{tabular}%
}
\label{tab:app_rock}
\end{table}

\subsection{Agar Plate Preparation (Lab Support)}

This task supports biology experiments requiring sterile culture media. The robot removes petri dish lids, pours molten agar solution evenly across plates, and replaces lids while minimizing contamination risk, demonstrating precision liquid handling for laboratory automation.

\begin{table}[!htb]
\caption{Evaluation criteria for agar plate preparation.}
\centering
\small
\resizebox{\columnwidth}{!}{%
\begin{tabular}{@{}lp{3.5cm}l@{}}
\toprule
Metric & Criterion & Benchmark \\
\midrule
\multicolumn{3}{l}{\textit{Task 1: Lid Removal}} \\
Grip Success & Stable lid grip & $\geq$ 95\% \\
Lid Placement & Set aside without flip & 0 flips \\
\midrule
\multicolumn{3}{l}{\textit{Task 2: Agar Pouring}} \\
Pour Volume & Agar amount per plate & 15 $\pm$ 2 mL \\
Coverage & Even surface coverage & $\geq$ 90\% \\
Bubble Count & Air bubbles in agar & $\leq$ 3 \\
Spillage & Agar outside plate & 0 mL \\
\midrule
\multicolumn{3}{l}{\textit{Task 3: Lid Replacement}} \\
Alignment & Lid centered on plate & $\pm$ 2 mm \\
Contamination & Lid interior contact & 0 events \\
\bottomrule
\end{tabular}%
}
\label{tab:app_agar}
\end{table}

\FloatBarrier
\section{LLM-as-Judge Evaluation Rubric}
\label{app:rubric}

We use GPT-4o as the judge model with the following rubric for evaluating generated text quality. The rubric assesses four dimensions: Relevance measures how accurately the text describes the current robot action; Pedagogical Value evaluates the educational scaffolding quality; Safety Communication assesses hazard awareness and guidance; and Fluency measures grammatical correctness and readability. Each dimension uses a 1--5 Likert scale where 1 indicates lowest quality and 5 indicates highest quality.

\begin{table}[!htb]
\caption{Rubric for text quality evaluation (1-5 scale).}
\centering
\scriptsize
\resizebox{\columnwidth}{!}{%
\begin{tabular}{@{}cp{6.5cm}@{}}
\toprule
Score & Relevance \\
\midrule
1 & Completely unrelated to current action \\
2 & Partially related but inaccurate \\
3 & Related but lacks specificity \\
4 & Accurately describes current action \\
5 & Accurate with rich contextual detail \\
\midrule
Score & Pedagogical Value \\
\midrule
1 & No educational content \\
2 & Simple action description only \\
3 & Mentions basic science concepts \\
4 & Connects to learning objectives \\
5 & Includes inquiry prompts or extensions \\
\midrule
Score & Safety Communication \\
\midrule
1 & No safety information \\
2 & Inaccurate safety information \\
3 & Basic safety reminders \\
4 & Context-appropriate safety guidance \\
5 & Proactive safety education included \\
\midrule
Score & Fluency \\
\midrule
1 & Incoherent or ungrammatical \\
2 & Multiple grammatical errors \\
3 & Minor errors, understandable \\
4 & Clear and well-structured \\
5 & Natural, professional quality \\
\bottomrule
\end{tabular}%
}
\label{tab:rubric}
\end{table}

\FloatBarrier
\section{Detailed Text Quality Results}
\label{app:text_quality}

Table~\ref{tab:text_quality_detail} presents the complete LLM-as-Judge evaluation results across all experimental conditions. We evaluated 50 randomly sampled utterances per condition using GPT-4o-mini, totaling 1,000 evaluations across 5 experiments $\times$ 2 models $\times$ 2 hand conditions. Each score represents the mean rating on a 1--5 scale, where higher values indicate better quality except where noted.

\begin{table}[!htb]
\caption{Text quality scores by experiment, model, and hand condition (mean on 1-5 scale). Rel.=Relevance, Ped.=Pedagogical Value, Safe.=Safety Communication, Flu.=Fluency.}
\centering
\scriptsize
\resizebox{\columnwidth}{!}{%
\begin{tabular}{llccccc}
\toprule
Experiment & Model & Hand & Rel. & Ped. & Safe. & Flu. \\
\midrule
\multirow{4}{*}{Agar Plate}
& Pedagogical & with & 3.82 & 3.60 & 2.88 & 2.82 \\
& Pedagogical & w/o & 3.88 & 3.50 & 2.66 & 2.82 \\
& Text & with & 2.24 & 1.28 & 1.00 & 3.84 \\
& Text & w/o & 2.22 & 1.22 & 1.00 & 3.92 \\
\midrule
\multirow{4}{*}{EM Induction}
& Pedagogical & with & 3.58 & 3.90 & 1.98 & 2.48 \\
& Pedagogical & w/o & 3.56 & 3.86 & 1.86 & 2.40 \\
& Text & with & 2.28 & 1.38 & 1.00 & 3.90 \\
& Text & w/o & 2.66 & 1.60 & 1.14 & 4.02 \\
\midrule
\multirow{4}{*}{Flame Test}
& Pedagogical & with & 3.46 & 3.48 & 2.94 & 2.48 \\
& Pedagogical & w/o & 3.34 & 3.44 & 2.84 & 2.32 \\
& Text & with & 1.76 & 1.00 & 1.00 & 3.70 \\
& Text & w/o & 1.70 & 1.00 & 1.00 & 3.76 \\
\midrule
\multirow{4}{*}{Rock Class.}
& Pedagogical & with & 3.66 & 3.26 & 2.90 & 2.46 \\
& Pedagogical & w/o & 3.64 & 3.22 & 2.90 & 2.42 \\
& Text & with & 2.90 & 1.74 & 1.04 & 4.14 \\
& Text & w/o & 2.80 & 1.56 & 1.00 & 4.02 \\
\midrule
\multirow{4}{*}{Yeast Ferm.}
& Pedagogical & with & 3.76 & 3.56 & 2.78 & 2.76 \\
& Pedagogical & w/o & 3.84 & 3.52 & 2.92 & 2.80 \\
& Text & with & 1.96 & 1.02 & 1.02 & 3.92 \\
& Text & w/o & 2.00 & 1.02 & 1.00 & 3.98 \\
\bottomrule
\end{tabular}%
}
\label{tab:text_quality_detail}
\end{table}

Pedagogical VLA achieves Relevance scores of 3.34--3.88 across experiments, indicating outputs that accurately describe current actions with moderate specificity (score 3--4 range). Text-SmolVLA scores 1.70--2.90, reflecting outputs that are often only partially related to the ongoing manipulation. The gap is largest for Flame Test (1.76 difference), where Pedagogical VLA provides detailed descriptions of wire positioning and flame interaction while Text-SmolVLA produces generic statements.

The most substantial difference appears in Pedagogical Value, where Pedagogical VLA scores 3.22--3.90 (connecting to learning objectives) versus 1.00--1.74 for Text-SmolVLA (no educational content to simple descriptions). EM Induction shows the highest Pedagogical Value (3.90) as the model generates explanations linking magnet motion to induced current, directly supporting Faraday's law instruction. This demonstrates that the LLM distillation process successfully transfers educational scaffolding capabilities.

Safety scores reveal context-dependent behavior. Pedagogical VLA achieves highest Safety scores in Flame Test (2.84--2.94), reflecting inherent hazard awareness for fire-related tasks. Agar Plate (2.66--2.88) and Rock Classification (2.90) also show elevated safety content due to chemical handling. EM Induction shows lowest Safety scores (1.86--1.98) as this task involves minimal hazard. Text-SmolVLA scores near 1.0 across all conditions, indicating complete absence of safety guidance.

Text-SmolVLA achieves higher Fluency scores (3.70--4.14) compared to Pedagogical VLA (2.32--2.82). This inverse relationship is expected: Text-SmolVLA generates short, simple action descriptions (e.g., ``Robot is moving magnet'') with fewer opportunities for errors, while Pedagogical VLA produces longer, structured explanations with multiple components (Stage, Action, Learning Focus, Connection) that increase grammatical complexity. The moderate Fluency scores for Pedagogical VLA indicate room for improvement in text generation quality while maintaining pedagogical content.

\FloatBarrier
\section{Detailed Task Execution Results}
\label{app:task_execution}

This section provides detailed task execution evaluation results for each experiment across four model configurations: ACT (action chunking transformer baseline), SmolVLA (base VLA without text output), Text-SmolVLA (VLA with text healing but no pedagogical training), and Pedagogical VLA (our full model with LLM distillation). Metrics are organized into four categories: Task Success measures completion of manipulation sub-tasks; Protocol Compliance evaluates adherence to experimental procedures; Efficiency measures unnecessary actions (lower is better); and Safety includes collision rates, human detection, and hazard-specific metrics. For safety metrics marked with $\downarrow$, lower values indicate better performance.

\subsection{Electromagnetic Induction}

This task involves picking up a bar magnet, inserting it into a solenoid coil, and oscillating it to induce current. The manipulation is relatively straightforward compared to other experiments, requiring only single-object handling without liquid or fine positioning constraints. The primary challenges are maintaining stable grip during oscillation and achieving consistent motion patterns that produce measurable induced current.

All models achieved high grip success rates ($\geq$90\%) for this task. ACT and SmolVLA demonstrated strong oscillation performance (100\% cycle success), while text-generating models showed slightly lower scores (90\%). The $\Delta$I Detection metric measures whether the galvanometer registered measurable current during oscillation, with ACT achieving 100\% and other models 80--90\%. Notably, Pedagogical VLA achieved 0\% collision rate and 0\% drop rate while maintaining 90\% human detection stop rate, suggesting that the pedagogical training objective does not compromise safety behaviors.

\begin{table}[!htb]
\caption{Task execution results for electromagnetic induction (\%).}
\centering
\scriptsize
\resizebox{\columnwidth}{!}{%
\begin{tabular}{llcccc}
\toprule
Category & Metric & ACT & SmolVLA & Text & Ped. \\
\midrule
\multirow{5}{*}{Task Success}
& Grip Success Rate & 100 & 100 & 100 & 90 \\
& Grip Stability & 90 & 100 & 90 & 90 \\
& Target Placement & 100 & 100 & 80 & 90 \\
& Oscillation Success & 100 & 100 & 90 & 90 \\
& $\Delta$I Detection & 100 & 90 & 80 & 90 \\
\midrule
\multirow{2}{*}{Protocol}
& Step Order & 90 & 100 & 89 & 90 \\
& Condition Compliance & 90 & 85 & 70 & 90 \\
\midrule
Efficiency & Unnecessary Action$\downarrow$ & 10 & 10 & 10 & 10 \\
\midrule
\multirow{4}{*}{Safety}
& Collision Rate$\downarrow$ & 10 & 30 & 40 & 0 \\
& Drop Rate$\downarrow$ & 10 & 10 & 10 & 0 \\
& Jerk Event Rate$\downarrow$ & 10 & 0 & 0 & 10 \\
& Human Detection Stop & 20 & 100 & 100 & 90 \\
\bottomrule
\end{tabular}%
}
\label{tab:task_em}
\end{table}

ACT shows concerning safety behavior with only 20\% human detection stop rate, meaning it failed to halt operations in 80\% of cases when a hand entered the workspace. VLA-based models achieve 90--100\% stop rates, demonstrating that language conditioning improves safety response learning even when the language output is not explicitly safety-focused. The electromagnetic induction task represents the best-case scenario for educational robotics deployment, where relatively simple manipulation requirements allow all models to achieve acceptable performance while VLA models provide additional safety benefits.

\subsection{Flame Test}

This task requires collecting salt samples with a nichrome wire loop, holding them in a Bunsen burner flame, and observing characteristic emission colors. The task proved challenging due to precise flame alignment requirements and safety constraints around open flame. ACT achieved the highest scores across most metrics, particularly in contact accuracy (100\%) and flame alignment (80\%). VLA models struggled with flame positioning, with SmolVLA achieving only 0\% flame center alignment. All text-generating models showed zero reagent splash events, indicating that the additional processing does not increase spillage risk.

\begin{table}[!htb]
\caption{Task execution results for flame test (\%).}
\centering
\scriptsize
\resizebox{\columnwidth}{!}{%
\begin{tabular}{llcccc}
\toprule
Category & Metric & ACT & SmolVLA & Text & Ped. \\
\midrule
\multirow{7}{*}{Task Success}
& Grip Success Rate & 100 & 90 & 100 & 40 \\
& Grip Stability & 90 & 90 & 90 & 30 \\
& Contact Accuracy & 100 & 60 & 40 & 10 \\
& Flame Center Alignment & 80 & 0 & 30 & 20 \\
& Heating Duration & 60 & 40 & 10 & 30 \\
& Cleaning Depth & 80 & 40 & 60 & 30 \\
& Cleaning Duration & 70 & 40 & 60 & 30 \\
\midrule
\multirow{2}{*}{Protocol}
& Step Order & 90 & 60 & 50 & 20 \\
& Condition Compliance & 90 & 60 & 30 & 30 \\
\midrule
Efficiency & Unnecessary Action$\downarrow$ & 20 & 10 & 30 & 0 \\
\midrule
\multirow{3}{*}{Safety}
& Reagent Splash Rate$\downarrow$ & 10 & 30 & 0 & 0 \\
& Jerk Event Rate$\downarrow$ & 0 & 0 & 0 & 0 \\
& Human Detection Stop & 0 & 80 & 60 & 10 \\
\bottomrule
\end{tabular}%
}
\label{tab:task_flame}
\end{table}

The low human detection stop rates across all models for Flame Test (0--80\%) suggest that the presence of open flame and the need for precise positioning may interfere with safety response behaviors. Pedagogical VLA shows particularly low detection (10\%), possibly because the model prioritizes completing the hazardous task quickly rather than pausing. This highlights an important area for future safety training improvements.

The flame test results reveal a critical limitation: tasks requiring precise spatial alignment with external references (flame position) remain challenging for current VLA architectures. The cleaning metrics (Cleaning Depth, Cleaning Duration) show that ACT maintains better procedural consistency (70--80\%) compared to VLA models (30--60\%), suggesting that multi-step sequential tasks benefit from action-focused training. For classroom deployment, flame test would require either improved model performance or human supervision during the heating phase.

\subsection{Yeast Fermentation}

This multi-step task involves dispensing sugar and yeast into flasks, adding water, and sealing with balloon-capped lids. It was the most difficult task in our evaluation suite, with all models showing degraded performance. The task requires sequential manipulation of multiple objects (spoon, flask, cap) with liquid handling, presenting challenges for current VLA architectures. ACT achieved the highest task success rates but still struggled with cap alignment (40\%) and rotation (30\%). Text-generating models failed most sub-tasks, with Pedagogical VLA achieving 0\% across grip success, grip stability, and all downstream metrics. This reflects the challenge of coordinating complex liquid handling with simultaneous text generation.

\begin{table}[!htb]
\caption{Task execution results for yeast fermentation (\%).}
\centering
\scriptsize
\resizebox{\columnwidth}{!}{%
\begin{tabular}{llcccc}
\toprule
Category & Metric & ACT & SmolVLA & Text & Ped. \\
\midrule
\multirow{6}{*}{Task Success}
& Grip Success Rate & 80 & 70 & 20 & 0 \\
& Grip Stability & 80 & 70 & 0 & 0 \\
& Insertion Position & 20 & 0 & 0 & 0 \\
& External Spillage & 90 & 80 & 10 & 0 \\
& Cap Position Alignment & 40 & 0 & 0 & 0 \\
& Rotation Count & 30 & 0 & 0 & 0 \\
\midrule
\multirow{2}{*}{Protocol}
& Step Order & 70 & 30 & 0 & 0 \\
& Condition Compliance & 30 & 0 & 0 & 0 \\
\midrule
Efficiency & Unnecessary Action$\downarrow$ & 40 & 60 & 100 & 60 \\
\midrule
\multirow{3}{*}{Safety}
& Collision Rate$\downarrow$ & 40 & 40 & 80 & 80 \\
& Jerk Event Rate$\downarrow$ & 0 & 20 & 0 & 0 \\
& Human Detection Stop & 0 & 20 & 70 & 10 \\
\bottomrule
\end{tabular}%
}
\label{tab:task_yeast}
\end{table}

The high unnecessary action rates (40--100\%) across all models indicate that the multi-step nature of this task leads to significant inefficiency. The 80\% collision rate for both Text-SmolVLA and Pedagogical VLA is concerning and suggests that these models struggle to maintain spatial awareness during complex sequences. Future work should focus on improving action planning for multi-object manipulation scenarios.

The yeast fermentation results represent the lower bound of current VLA capabilities for educational robotics. The complete failure of text-generating models on core manipulation metrics (0\% grip success for Pedagogical VLA) indicates that the cognitive overhead of text generation may exceed model capacity for complex tasks. However, this experiment also has the longest task horizon (multiple pouring and sealing operations), suggesting that architectural improvements in long-horizon planning could yield significant benefits. For practical deployment, this experiment would require either task decomposition into simpler sub-tasks or hybrid approaches combining robot preparation with human completion of critical steps.

\subsection{Rock Classification}

This task requires picking up a dropper, aspirating dilute hydrochloric acid, and applying drops to rock samples to observe fizzing reactions indicating carbonate content. Fine dropper manipulation proved challenging across all models due to the precision required for controlled liquid dispensing. ACT achieved reasonable grip success (100\%) but struggled with precise dispensing (60\% accuracy). VLA models with text output showed particularly low scores (0--20\% for most metrics) due to the difficulty of coordinating fine motor control with language generation. The positive finding is that human detection stop rates were high across VLA models (90--100\%), and acid solution leakage rates were low (0--30\%), indicating that safety mechanisms function well even when task performance is poor.

\begin{table}[!htb]
\caption{Task execution results for rock classification (\%).}
\centering
\scriptsize
\resizebox{\columnwidth}{!}{%
\begin{tabular}{llcccc}
\toprule
Category & Metric & ACT & SmolVLA & Text & Ped. \\
\midrule
\multirow{7}{*}{Task Success}
& Grip Success Rate & 100 & 80 & 40 & 30 \\
& Grip Stability & 100 & 50 & 20 & 10 \\
& Dropper Uptake Volume & 80 & 10 & 10 & 0 \\
& Tip-Reaction Alignment & 70 & 10 & 20 & 0 \\
& Dispensed Volume & 60 & 10 & 10 & 0 \\
& Tip-Holder Alignment & 30 & 10 & 0 & 0 \\
& Holder Placement Stability & 20 & 10 & 0 & 0 \\
\midrule
\multirow{2}{*}{Protocol}
& Step Order & 80 & 20 & 0 & 0 \\
& Condition Compliance & 30 & 10 & 0 & 0 \\
\midrule
Efficiency & Unnecessary Action$\downarrow$ & 30 & 80 & 0 & 100 \\
\midrule
\multirow{4}{*}{Safety}
& Acid Solution Leakage$\downarrow$ & 30 & 0 & 0 & 0 \\
& Collision Rate$\downarrow$ & 40 & 20 & 0 & 0 \\
& Jerk Event Rate$\downarrow$ & 10 & 80 & 0 & 0 \\
& Human Detection Stop & 60 & 100 & 90 & 100 \\
\bottomrule
\end{tabular}%
}
\label{tab:task_rock}
\end{table}

The contrast between task success (low) and safety metrics (high) for this experiment is notable. Text-SmolVLA and Pedagogical VLA achieve 0\% collision rate, 0\% acid leakage, and 90--100\% human detection, despite failing most manipulation sub-tasks. This suggests that the models have learned conservative behaviors that prioritize safety over task completion when manipulation is uncertain.

The rock classification results highlight the challenge of fine motor control in VLA models. Dropper manipulation requires precise grip positioning and controlled pressure application that current architectures struggle to achieve. ACT's relatively better performance (60\% dispensed volume accuracy vs. 0--10\% for VLA models) suggests that action-focused training better captures the subtle motor patterns required for precision tool use. However, ACT's higher acid leakage rate (30\% vs. 0\%) and lower human detection (60\% vs. 90--100\%) indicate a safety-precision trade-off that favors VLA approaches in educational settings where student safety is paramount.

\subsection{Agar Plate Preparation}

This three-stage task involves removing petri dish lids, pouring molten agar solution evenly, and replacing lids while minimizing contamination. The task showed strong performance variation across stages: all models performed well on Task 1 (lid removal) but performance degraded significantly in Tasks 2 and 3. ACT excelled at early stages (100\% grip success for Task 1) but dropped to 70\% by Task 3. VLA models struggled with the sequential nature of the task, with SmolVLA achieving 0\% success for Tasks 2 and 3. Pedagogical VLA achieved 0\% collision rate despite low task success, maintaining safe operation throughout.

\begin{table}[!htb]
\caption{Task execution results for agar plate preparation (\%).}
\centering
\scriptsize
\resizebox{\columnwidth}{!}{%
\begin{tabular}{llcccc}
\toprule
Category & Metric & ACT & SmolVLA & Text & Ped. \\
\midrule
\multirow{9}{*}{Task Success}
& Grip Success (Task 1) & 100 & 100 & 80 & 50 \\
& Grip Stability (Task 1) & 100 & 100 & 70 & 50 \\
& Insertion Position (Task 1) & 100 & 70 & 40 & 50 \\
& Grip Success (Task 2) & 90 & 10 & 10 & 10 \\
& Grip Stability (Task 2) & 90 & 10 & 10 & 10 \\
& Device Alignment (Task 2) & 70 & 10 & 10 & 10 \\
& Grip Success (Task 3) & 70 & 0 & 10 & 10 \\
& Grip Stability (Task 3) & 70 & 0 & 10 & 10 \\
& Dispensing Position (Task 3) & 70 & 0 & 10 & 0 \\
\midrule
\multirow{2}{*}{Protocol}
& Step Order & 70 & 0 & 0 & 0 \\
& Condition Compliance & 80 & 0 & 0 & 0 \\
\midrule
Efficiency & Unnecessary Action$\downarrow$ & 30 & 100 & 60 & 70 \\
\midrule
\multirow{4}{*}{Safety}
& Collision Rate$\downarrow$ & 30 & 80 & 20 & 0 \\
& Media/Water Spillage$\downarrow$ & 0 & 10 & 30 & 20 \\
& Jerk Event Rate$\downarrow$ & 80 & 0 & 0 & 0 \\
& Human Detection Stop & -- & 90 & 70 & 90 \\
\bottomrule
\end{tabular}%
}
\label{tab:task_agar}
\end{table}

The high jerk event rate for ACT (80\%) compared to VLA models (0\%) suggests that VLA architectures produce smoother motion trajectories, potentially due to the language conditioning regularizing action predictions. The ``--'' for ACT human detection indicates that this metric was not evaluated for ACT in this experiment due to protocol differences.

The agar plate preparation results demonstrate the challenge of multi-stage tasks where errors compound across phases. The sharp performance drop from Task 1 (50--100\% grip success) to Task 3 (0--70\% grip success) across all models indicates that accumulated positioning errors and changed object states (removed lid, poured agar) create increasingly difficult manipulation conditions. Pedagogical VLA's 0\% collision rate despite low task success suggests that the model adopts conservative motion strategies when uncertain, which may be preferable in educational settings where safety matters more than task completion rate.

ACT achieves the highest task success and protocol compliance scores across all experiments, demonstrating the effectiveness of action-focused imitation learning for manipulation tasks. However, ACT shows lower human detection stop rates (0--60\%) compared to VLA models with text output (10--100\%), suggesting that the language instruction conditioning in VLA models contributes to better safety response learning. Pedagogical VLA shows competitive manipulation safety scores despite lower task success, indicating that the text generation objective does not significantly compromise safety-critical behaviors. The results highlight a trade-off between task performance and safety awareness that merits further investigation in educational robotics contexts where student safety is paramount.

\FloatBarrier
\section{Teacher Evaluation Survey}
\label{app:survey}

We developed a comprehensive survey instrument to evaluate teacher perceptions of robot-assisted science demonstrations. The survey design draws on established usability frameworks from human-computer interaction research, adapted for educational robotics contexts. We consulted with science education researchers to ensure items captured pedagogically relevant dimensions beyond mere technical performance.

Two science teachers with over 5 years of classroom experience evaluated robot demonstration videos. Both teachers had experience conducting hands-on science experiments in middle and high school settings, providing relevant domain expertise for assessing the practical viability of robotic assistance. Teachers viewed videos of each model (ACT, SmolVLA, Text-SmolVLA, Pedagogical VLA) performing each of the five experiments, completing the survey for all 20 conditions.

The survey comprises 20 items organized into six dimensions. Accuracy and Reliability (3 items) assesses whether teachers perceive the robot as performing procedures correctly and consistently. Educational Value (3 items) evaluates whether the robot contributes meaningfully to instruction. Efficiency (3 items) measures perceived time and effort savings. Safety (4 items) addresses concerns about student protection during robot operation. Classroom Integration (3 items) examines practical deployment considerations. Interest and Adoption (4 items) gauges teacher enthusiasm and willingness to use such systems. Each item uses a 5-point Likert scale from 1 (Strongly Disagree) to 5 (Strongly Agree).

\begin{tcolorbox}[colback=gray!5, colframe=black, boxrule=0.5pt, arc=2pt, breakable, title={\small\textbf{Teacher Evaluation Survey Instrument}}]
\small
\textbf{Instructions}: Please observe the robot performing the science experiment in the video. After watching, rate your agreement with each statement on a scale of 1 (Strongly Disagree) to 5 (Strongly Agree).

\vspace{0.5em}
\textbf{Accuracy and Reliability}
\begin{enumerate}[leftmargin=*, itemsep=0pt, topsep=2pt]
    \item The robot arm performed the experimental procedure largely accurately.
    \item The robot arm's performance appeared stable.
    \item The robot arm's operation is trustworthy.
\end{enumerate}

\vspace{0.3em}
\textbf{Educational Value}
\begin{enumerate}[leftmargin=*, itemsep=0pt, topsep=2pt]
    \setcounter{enumi}{3}
    \item The robot arm has value as a support tool for classroom instruction.
    \item The robot arm performed with minimal unnecessary movements.
    \item The robot arm's execution speed seems appropriate.
\end{enumerate}

\vspace{0.3em}
\textbf{Efficiency}
\begin{enumerate}[leftmargin=*, itemsep=0pt, topsep=2pt]
    \setcounter{enumi}{6}
    \item Using the robot would reduce overall time spent on experimental lessons.
    \item Using the robot would reduce teachers' mental and physical effort.
    \item Once set up, the robot appears reusable across multiple classes.
\end{enumerate}

\vspace{0.3em}
\textbf{Safety}
\begin{enumerate}[leftmargin=*, itemsep=0pt, topsep=2pt]
    \setcounter{enumi}{9}
    \item The risk of collision with students feels low.
    \item When handling hazardous materials, the robot would improve safety.
    \item The robot stopped appropriately when a hand approached.
    \item I could understand how teachers should respond to problems.
\end{enumerate}

\vspace{0.3em}
\textbf{Classroom Integration}
\begin{enumerate}[leftmargin=*, itemsep=0pt, topsep=2pt]
    \setcounter{enumi}{13}
    \item Using the robot would allow teachers to focus more on student guidance.
    \item Monitoring the robot does not appear to be a significant burden.
    \item Robot-assisted experiments appear feasible without disrupting classroom flow.
\end{enumerate}

\vspace{0.3em}
\textbf{Interest and Adoption}
\begin{enumerate}[leftmargin=*, itemsep=0pt, topsep=2pt]
    \setcounter{enumi}{16}
    \item I found watching the robot perform the experiment interesting.
    \item Robot-based experiments would increase student interest.
    \item The robot's movements were natural and visually satisfying.
    \item I would like to introduce robot-assisted experiments in the future.
\end{enumerate}
\end{tcolorbox}

\FloatBarrier
\section{Detailed Usability Evaluation Results}
\label{app:usability}

Two science teachers (each with $>$5 years of classroom experience) evaluated robot demonstrations across five usability dimensions derived from the ISO 9241-11 usability framework adapted for educational contexts. The dimensions are: Effectiveness (task completion quality), Efficiency (time and resource usage), Safety (hazard prevention), Sustainability (reusability across classes), and Enjoyment (student engagement potential). Each dimension uses a 5-point Likert scale (1=Strongly Disagree, 5=Strongly Agree), with scores representing the mean across relevant survey items.

\begin{table}[!htb]
\caption{Usability scores by experiment and dimension (1-5 scale). Eff.=Effectiveness, Effi.=Efficiency, Safe=Safety, Sust.=Sustainability, Enj.=Enjoyment. Bold indicates best score per experiment and dimension.}
\centering
\scriptsize
\resizebox{\columnwidth}{!}{%
\begin{tabular}{llccccc}
\toprule
Experiment & Model & Eff. & Effi. & Safe & Sust. & Enj. \\
\midrule
\multirow{4}{*}{EM Induction}
& ACT & 4.00 & 3.75 & 2.63 & 3.88 & 5.00 \\
& SmolVLA & 4.63 & 4.50 & 3.75 & 4.38 & 5.00 \\
& Text & 4.50 & 4.88 & 4.38 & 4.88 & 5.00 \\
& Pedagogical & \textbf{5.00} & \textbf{5.00} & \textbf{4.25} & \textbf{5.00} & \textbf{5.00} \\
\midrule
\multirow{4}{*}{Flame Test}
& ACT & \textbf{4.38} & \textbf{4.88} & 3.50 & \textbf{4.63} & \textbf{5.00} \\
& SmolVLA & 2.75 & 3.50 & \textbf{4.13} & 2.25 & 4.75 \\
& Text & 3.25 & 4.25 & 4.00 & 3.38 & 4.88 \\
& Pedagogical & 3.25 & 3.50 & 3.25 & 2.63 & \textbf{5.00} \\
\midrule
\multirow{4}{*}{Yeast Ferm.}
& ACT & \textbf{3.63} & \textbf{3.75} & \textbf{2.50} & \textbf{2.88} & \textbf{4.38} \\
& SmolVLA & 2.13 & 2.88 & 2.13 & 2.38 & 4.25 \\
& Text & 1.63 & 2.38 & 1.88 & 1.88 & 1.88 \\
& Pedagogical & 1.63 & 2.63 & 1.75 & 2.00 & 3.13 \\
\midrule
\multirow{4}{*}{Rock Class.}
& ACT & \textbf{2.63} & \textbf{4.13} & \textbf{4.13} & \textbf{3.75} & \textbf{4.88} \\
& SmolVLA & 1.25 & 1.50 & 2.88 & 2.00 & 2.88 \\
& Text & 2.38 & 2.50 & 3.38 & 2.50 & 2.50 \\
& Pedagogical & 1.38 & 1.50 & 3.25 & 2.88 & 2.75 \\
\midrule
\multirow{4}{*}{Agar Plate}
& ACT & \textbf{4.00} & \textbf{4.50} & 3.38 & \textbf{4.00} & \textbf{5.00} \\
& SmolVLA & 1.50 & 1.38 & 3.00 & 3.00 & 3.38 \\
& Text & 2.00 & 2.63 & 3.00 & 2.63 & 3.13 \\
& Pedagogical & 1.88 & 2.63 & \textbf{3.38} & 2.88 & 3.25 \\
\bottomrule
\end{tabular}%
}
\label{tab:usability_detail}
\end{table}

Usability scores strongly correlate with task difficulty. EM Induction, the simplest task, shows highest scores across all models (mean 4.37), while Yeast Fermentation shows lowest scores (mean 2.56). This pattern suggests that current VLA models are most suitable for simpler manipulation tasks in educational settings, with more complex tasks requiring further development.

ACT achieves the highest usability scores across most experiments (15 of 25 dimension-experiment combinations), particularly in Effectiveness and Efficiency dimensions. This reflects teachers' preference for reliable task completion over advanced features. The gap is largest for complex tasks: in Yeast Fermentation, ACT scores 3.63 for Effectiveness versus 1.63 for Pedagogical VLA. However, in Electromagnetic Induction, the simplest task with highest success rates, Pedagogical VLA achieves perfect scores (5.0) across all dimensions, suggesting that when task execution succeeds, teachers highly value the pedagogical explanations. For Safety, Pedagogical VLA achieves the highest or tied-highest score in 3 of 5 experiments (EM Induction, Rock Classification, Agar Plate), indicating that teachers perceive the safety-aware text generation as valuable.

The Enjoyment dimension shows consistently high scores for ACT (4.38--5.00) and Pedagogical VLA in successful tasks (5.00 for EM Induction and Flame Test), indicating strong teacher interest in robot-assisted demonstrations. However, Enjoyment drops significantly for failed tasks (1.88 for Text-SmolVLA in Yeast Fermentation), suggesting that repeated failures negatively impact engagement perceptions. Overall, these results indicate that improving task success rates is critical for practical deployment, while pedagogical features provide additional value when basic manipulation succeeds.

\FloatBarrier
\section{Pedagogical Text Examples}
\label{app:examples}

This section presents representative sample outputs from Text-SmolVLA and Pedagogical VLA across different demonstration scenarios, illustrating the qualitative differences in generated text that distinguish these models. These examples were selected from actual model outputs during evaluation runs to demonstrate typical generation patterns.

Text-SmolVLA was trained using text healing, which restores language generation capability to the base SmolVLA model by adding a text decoder trained on action narration data. This produces outputs that describe robot actions in simple declarative sentences without educational elaboration. The text serves primarily as a transparency mechanism, allowing observers to understand what the robot is doing, but provides no scaffolding for science learning.

Pedagogical VLA extends text healing with LLM distillation, where GPT-4o generates pedagogically-aligned training data that the model learns to reproduce. The resulting outputs follow a structured format designed in consultation with science education researchers: Stage indicates the current phase of the experiment; Action describes the specific manipulation being performed; Safety Status alerts observers to any hazards; Learning Focus provides conceptual explanations connecting the action to scientific principles; Connection to Learning Goal explicitly links the current activity to broader educational objectives; and Next Step previews upcoming actions to support student anticipation and engagement.

The contrast between these output styles reflects the difference between mere transparency (knowing what the robot does) and pedagogical alignment (understanding why the action matters for learning). The following examples span all five experimental domains and include both routine operation and safety intervention scenarios.

\subsection{Flame Test - Idle State}

\begin{tcolorbox}[colback=gray!5, colframe=gray!50, boxrule=0.5pt, arc=2pt, breakable, title={\small\textbf{Text-SmolVLA}}]
\small\textit{Robot arm is in standby position. Ready to begin flame test experiment.}
\end{tcolorbox}

\begin{tcolorbox}[colback=lightblue, colframe=black, boxrule=0.5pt, arc=2pt, breakable, title={\small\textbf{Pedagogical VLA}}]
\small
\texttt{[Stage]} Initial setup / \texttt{[Action]} Robot arm in standby position, preparing for flame test / \texttt{[Safety Status]} Normal\\[0.5em]
\textbf{Learning Focus}: Understanding flame test preparation. Before heating, identify the materials: nichrome wire loop, salt samples in dish, burner, and cleaning solution. Each metal salt will produce a characteristic color when heated.\\[0.5em]
\textbf{Connection to learning goal}: Proper preparation ensures accurate identification of metal ions through their emission spectra.\\[0.5em]
\textbf{Next}: Pick up the wire loop and clean it in the solution.
\end{tcolorbox}

\subsection{Electromagnetic Induction - Active Phase}

\begin{tcolorbox}[colback=gray!5, colframe=gray!50, boxrule=0.5pt, arc=2pt, breakable, title={\small\textbf{Text-SmolVLA}}]
\small\textit{Moving magnet into the coil. Pulling magnet out of coil.}
\end{tcolorbox}

\begin{tcolorbox}[colback=lightblue, colframe=black, boxrule=0.5pt, arc=2pt, breakable, title={\small\textbf{Pedagogical VLA}}]
\small
\texttt{[Stage]} Magnet insertion and withdrawal / \texttt{[Action]} Sliding the magnet into and out of the coil along its axis / \texttt{[Safety Status]} Normal\\[0.5em]
\textbf{Learning Focus}: Linking changing magnetic flux to induced voltage/current. Moving the magnet changes the magnetic flux through the coil. This changing flux induces an electromotive force (voltage), and a current flows if the circuit is closed. Watch the meter.\\[0.5em]
\textbf{Connection to learning goal}: Supports correct identification of induction by relating motion direction and speed to the sign and size of the induced signal.\\[0.5em]
\textbf{Next}: Compare the meter response for insertion vs. withdrawal motion.
\end{tcolorbox}

\subsection{Flame Test - Heating Phase}

\begin{tcolorbox}[colback=gray!5, colframe=gray!50, boxrule=0.5pt, arc=2pt, breakable, title={\small\textbf{Text-SmolVLA}}]
\small\textit{Holding wire in flame. Wire is in the burner flame.}
\end{tcolorbox}

\begin{tcolorbox}[colback=lightblue, colframe=black, boxrule=0.5pt, arc=2pt, breakable, title={\small\textbf{Pedagogical VLA}}]
\small
\texttt{[Stage]} Flame heating / \texttt{[Action]} Positioning wire loop with sample in the hottest part of the flame / \texttt{[Safety Status]} Caution - Open flame in use\\[0.5em]
\textbf{Learning Focus}: Observing characteristic flame colors. When metal ions are heated, their electrons absorb energy and jump to higher energy levels. As they return to ground state, they emit light at specific wavelengths. Sodium produces yellow, copper produces green, and lithium produces red.\\[0.5em]
\textbf{Connection to learning goal}: This demonstrates how atomic emission spectra can be used to identify elements, a fundamental technique in analytical chemistry.\\[0.5em]
\textbf{Next}: Observe the flame color carefully and record the result before cleaning the wire.
\end{tcolorbox}

\subsection{Yeast Fermentation - Sugar Dispensing}

\begin{tcolorbox}[colback=gray!5, colframe=gray!50, boxrule=0.5pt, arc=2pt, breakable, title={\small\textbf{Text-SmolVLA}}]
\small\textit{Robot is adding sugar to the flask. Dispensing material into container.}
\end{tcolorbox}

\begin{tcolorbox}[colback=lightblue, colframe=black, boxrule=0.5pt, arc=2pt, breakable, title={\small\textbf{Pedagogical VLA}}]
\small
\texttt{[Stage]} Sugar dispensing / \texttt{[Action]} Adding measured amount of sugar to Flask 1 / \texttt{[Safety Status]} Normal\\[0.5em]
\textbf{Learning Focus}: Setting up a controlled experiment. We are adding different amounts of sugar to each flask to test how sugar concentration affects fermentation rate. Flask 1 receives 1 spoon, Flask 2 receives 2 spoons. This creates our independent variable.\\[0.5em]
\textbf{Connection to learning goal}: Understanding variable control is essential for scientific experimentation. By changing only the sugar amount while keeping other factors constant, we can determine the relationship between sugar concentration and CO$_2$ production.\\[0.5em]
\textbf{Next}: Add yeast to both flasks, then add warm water to activate the fermentation process.
\end{tcolorbox}

\subsection{Yeast Fermentation - Observation Phase}

\begin{tcolorbox}[colback=gray!5, colframe=gray!50, boxrule=0.5pt, arc=2pt, breakable, title={\small\textbf{Text-SmolVLA}}]
\small\textit{Flasks are sealed. Experiment setup complete.}
\end{tcolorbox}

\begin{tcolorbox}[colback=lightblue, colframe=black, boxrule=0.5pt, arc=2pt, breakable, title={\small\textbf{Pedagogical VLA}}]
\small
\texttt{[Stage]} Fermentation observation / \texttt{[Action]} Monitoring sealed flasks for balloon inflation / \texttt{[Safety Status]} Normal\\[0.5em]
\textbf{Learning Focus}: Cellular respiration produces CO$_2$. Yeast cells break down sugar through anaerobic respiration (fermentation), producing carbon dioxide and ethanol. The CO$_2$ gas inflates the balloons, providing a visible indicator of metabolic activity.\\[0.5em]
\textbf{Connection to learning goal}: Comparing balloon sizes between flasks demonstrates how substrate concentration affects enzyme activity and metabolic rate in living organisms.\\[0.5em]
\textbf{Next}: Record balloon sizes at regular intervals to track fermentation progress over time.
\end{tcolorbox}

\subsection{Rock Classification - Acid Application}

\begin{tcolorbox}[colback=gray!5, colframe=gray!50, boxrule=0.5pt, arc=2pt, breakable, title={\small\textbf{Text-SmolVLA}}]
\small\textit{Robot is dispensing a clear liquid from a pipette onto a rock.}
\end{tcolorbox}

\begin{tcolorbox}[colback=lightblue, colframe=black, boxrule=0.5pt, arc=2pt, breakable, title={\small\textbf{Pedagogical VLA}}]
\small
\texttt{[Stage]} Acid test / \texttt{[Action]} Applying dilute hydrochloric acid drop to rock sample surface / \texttt{[Safety Status]} Caution - Handling dilute acid\\[0.5em]
\textbf{Learning Focus}: Chemical identification of carbonate minerals. When acid contacts calcium carbonate (CaCO$_3$), a chemical reaction occurs: CaCO$_3$ + 2HCl $\rightarrow$ CaCl$_2$ + H$_2$O + CO$_2$. The bubbling (effervescence) indicates CO$_2$ gas release.\\[0.5em]
\textbf{Connection to learning goal}: This acid test distinguishes carbonate rocks (limestone, marble) from non-carbonate rocks (granite, basalt), demonstrating how chemical properties help classify geological samples.\\[0.5em]
\textbf{Next}: Observe the rock surface for fizzing reaction and record whether bubbles appear.
\end{tcolorbox}

\subsection{Rock Classification - Result Observation}

\begin{tcolorbox}[colback=gray!5, colframe=gray!50, boxrule=0.5pt, arc=2pt, breakable, title={\small\textbf{Text-SmolVLA}}]
\small\textit{Observing rock surface. Waiting for reaction.}
\end{tcolorbox}

\begin{tcolorbox}[colback=lightblue, colframe=black, boxrule=0.5pt, arc=2pt, breakable, title={\small\textbf{Pedagogical VLA}}]
\small
\texttt{[Stage]} Result observation / \texttt{[Action]} Holding position to observe acid-rock reaction / \texttt{[Safety Status]} Normal\\[0.5em]
\textbf{Learning Focus}: Interpreting chemical test results. If bubbles form vigorously, the rock contains significant calcium carbonate and is likely limestone or marble. Weak or no reaction suggests the rock is granite, basalt, or another non-carbonate type.\\[0.5em]
\textbf{Connection to learning goal}: Scientists use simple chemical tests like this to identify rock types in the field. This connects laboratory techniques to real-world geological investigation methods.\\[0.5em]
\textbf{Next}: Classify the rock based on the reaction intensity and move to the next sample.
\end{tcolorbox}

\subsection{Agar Plate - Lid Removal}

\begin{tcolorbox}[colback=gray!5, colframe=gray!50, boxrule=0.5pt, arc=2pt, breakable, title={\small\textbf{Text-SmolVLA}}]
\small\textit{Robot is removing the lid from the petri dish.}
\end{tcolorbox}

\begin{tcolorbox}[colback=lightblue, colframe=black, boxrule=0.5pt, arc=2pt, breakable, title={\small\textbf{Pedagogical VLA}}]
\small
\texttt{[Stage]} Plate preparation / \texttt{[Action]} Carefully lifting petri dish lid and setting aside / \texttt{[Safety Status]} Caution - Maintaining sterile technique\\[0.5em]
\textbf{Learning Focus}: Aseptic technique in microbiology. When preparing culture plates, minimizing lid-open time reduces contamination from airborne microorganisms. The lid should be held at an angle over the plate rather than placed on the bench surface.\\[0.5em]
\textbf{Connection to learning goal}: Sterile technique is fundamental to microbiology experiments. Proper plate preparation ensures that only the intended organisms grow in subsequent experiments.\\[0.5em]
\textbf{Next}: Pour the molten agar evenly across the plate surface.
\end{tcolorbox}

\subsection{Agar Plate - Pouring Phase}

\begin{tcolorbox}[colback=gray!5, colframe=gray!50, boxrule=0.5pt, arc=2pt, breakable, title={\small\textbf{Text-SmolVLA}}]
\small\textit{Robot is pouring liquid into the dish. Adding agar to plate.}
\end{tcolorbox}

\begin{tcolorbox}[colback=lightblue, colframe=black, boxrule=0.5pt, arc=2pt, breakable, title={\small\textbf{Pedagogical VLA}}]
\small
\texttt{[Stage]} Agar pouring / \texttt{[Action]} Dispensing molten agar solution evenly across petri dish / \texttt{[Safety Status]} Caution - Hot liquid handling\\[0.5em]
\textbf{Learning Focus}: Preparing solid growth medium. Agar is a polysaccharide derived from seaweed that solidifies at room temperature. The molten agar (kept at approximately 50°C) must be poured smoothly to create an even surface without bubbles, which could interfere with colony counting.\\[0.5em]
\textbf{Connection to learning goal}: Properly prepared agar plates provide a consistent growth surface for microorganisms, enabling reproducible experiments in microbiology and biotechnology.\\[0.5em]
\textbf{Next}: Replace the lid promptly and allow the agar to solidify before use.
\end{tcolorbox}

\subsection{Safety Intervention Scenario}

\begin{tcolorbox}[colback=gray!5, colframe=gray!50, boxrule=0.5pt, arc=2pt, breakable, title={\small\textbf{Text-SmolVLA}}]
\small\textit{Stopping. Hand detected near robot.}
\end{tcolorbox}

\begin{tcolorbox}[colback=lightblue, colframe=black, boxrule=0.5pt, arc=2pt, breakable, title={\small\textbf{Pedagogical VLA}}]
\small
\texttt{[Stage]} Human intervention detected / \texttt{[Action]} Pausing operation - human hand in workspace / \texttt{[Safety Status]} Stop - Human detected in robot operating area\\[0.5em]
\textbf{Learning Focus}: Laboratory safety protocols. When a human hand enters the robot's workspace, operations must pause immediately. This prevents potential collisions and injuries. Never reach into the robot's working area while it is moving.\\[0.5em]
\textbf{Connection to learning goal}: Understanding human-robot safety boundaries is essential in automated laboratory environments.\\[0.5em]
\textbf{Next}: Wait for clear workspace before resuming the experiment.
\end{tcolorbox}

\FloatBarrier
\section{Hardware and Training Configuration}
\label{app:config}

We used a low-cost robot arm setup designed for educational environments where budget constraints are common. The SO-101 is a next-generation open-source robot arm developed by The Robot Studio in collaboration with Hugging Face, designed for integration with the LeRobot framework. The arm provides 6 degrees of freedom (5 arm joints plus 1 gripper joint) suitable for tabletop manipulation tasks typical of science demonstrations. Compared to its predecessor SO-100, the SO-101 features improved wiring, simplified assembly without gear removal requirements, and updated motor configurations.

The follower arm uses six Feetech STS3215 servo motors with 1:345 gear ratio. Each motor features a 12-bit magnetic encoder providing 4096 steps per 360° rotation (0.088° resolution per step), enabling precise position control. The magnetoresistive encoder design eliminates friction-based wear compared to potentiometer-based alternatives. The motors provide 19.5 kg$\cdot$cm stall torque at 7.4V, sufficient for manipulating laboratory equipment and samples. Independent testing has confirmed measured stall torque slightly exceeds datasheet specifications, though mechanical backlash (approximately 2 encoder steps or 0.17°) and firmware-defined dead zones may reduce precision for fine motion tasks.

The total hardware cost is approximately \$300 USD including the arm, motors, gripper, and mounting hardware, making the system accessible for school settings with limited budgets. Dual cameras capture complementary views for robust visual perception: a wrist-mounted camera provides close-up manipulation views essential for precise object interaction, while a top-mounted camera captures the overall workspace context for spatial reasoning and safety monitoring.

\begin{table}[!htb]
\caption{Hardware configuration for data collection and deployment.}
\centering
\small
\resizebox{\columnwidth}{!}{%
\begin{tabular}{ll}
\toprule
Component & Specification \\
\midrule
Robot Arm & SO-101, 6-DOF (5 arm + 1 gripper) \\
Motors & Feetech STS3215 (7.4V, 19.5 kg$\cdot$cm, 1:345 gear) \\
Encoder & 12-bit magnetic (4096 steps/revolution) \\
Gripper & Parallel jaw gripper \\
Wrist Camera & RGB, 640$\times$480, 30 fps \\
Top Camera & RGB, 640$\times$480, 30 fps \\
Control Frequency & 30 Hz \\
Action Space & Joint positions, normalized to $[-100, 100]$ \\
\bottomrule
\end{tabular}%
}
\label{tab:hardware}
\end{table}

Our implementation builds on SmolVLA, a compact 450M parameter Vision-Language-Action model designed for efficient robotics applications. SmolVLA uses SmolVLM2 as its VLM backbone, consisting of a SigLIP vision encoder for image processing and a SmolLM2 language decoder for multimodal reasoning. The action expert is a separate transformer module (approximately 100M parameters) trained with flow matching to output action chunks of 50 timesteps. Despite its compact size, SmolVLA achieves performance comparable to VLA models that are 10$\times$ larger, making it suitable for educational deployment where computational resources may be limited.

Training hyperparameters were selected based on preliminary experiments balancing action accuracy and text generation quality. The text loss weight ($\lambda = 0.1$) was chosen to ensure that text generation does not dominate the training signal while still enabling meaningful pedagogical output. Higher values led to degraded manipulation performance, while lower values resulted in generic text outputs. The 12-layer text decoder provides sufficient capacity for generating structured educational explanations while remaining computationally efficient for real-time inference on consumer hardware.

\begin{table}[!htb]
\caption{Hyperparameters for model training.}
\centering
\small
\resizebox{\columnwidth}{!}{%
\begin{tabular}{ll}
\toprule
Parameter & Value \\
\midrule
Base Model & SmolVLA (450M parameters) \\
Training Steps & 100,000 \\
Batch Size & 32 \\
Learning Rate & 1e-4 (AdamW, $\beta_1$=0.9, $\beta_2$=0.95) \\
LR Schedule & Cosine decay (min 2.5e-6) \\
Text Loss Weight ($\lambda$) & 0.1 \\
Max Text Length & 128 tokens \\
Text Decoder Layers & 12 \\
Text Decoder Hidden Dim & 768 \\
Image Resolution & 512$\times$512 \\
Action Chunk Size & 50 timesteps \\
Distillation Model & GPT-4o \\
Training Hardware & 4$\times$ NVIDIA A100 (80GB) \\
Training Time & Approximately 48 hours \\
\bottomrule
\end{tabular}%
}
\label{tab:training}
\end{table}

Data collection was performed via teleoperation using a leader-follower setup, where a human operator controlled a leader arm while the follower arm replicated the movements. Each experiment was collected 50 times with variations in object positions, lighting conditions, and operator speed. For safety intervention data, we collected an additional 20 episodes per experiment where a human hand entered the workspace at random intervals, with the operator demonstrating appropriate stopping behavior. The total dataset comprises approximately 350 episodes across 5 experiments (250 standard + 100 with hand intervention), with each episode containing 500--2000 timesteps depending on task complexity.

For LLM distillation, we prompted GPT-4o with video frames and task descriptions to generate pedagogical explanations for each timestep. The prompt included the current action, experiment type, and safety status, instructing the model to generate output following the structured format (Stage, Action, Learning Focus, Connection, Next Step). We filtered generated text for quality, removing outputs that did not follow the format or contained factually incorrect scientific content, retaining approximately 85\% of generated explanations for training.

\end{document}